# Fractional-order Backpropagation Neural Networks Trained by Improved Fractional-order Steepest Descent Method

Yi-Fei PU and Jian WANG.

*Abstract*—This paper introduces a novel fractional-order branch of the family of BackPropagation Neural Networks (BPNNs) trained by the improved Fractional-order Steepest Descent Method (FSDM); this differs from the majority of the previous classic first-order BPNNs and as such trained by the traditional first-order steepest descent method. To improve the optimization performance of classic first-order BPNNs, in this paper we study whether it could be possible to apply improved FSDM based on fractional calculus to generalize classic first-order BPNNs to the Fractional-order Backpropagation Neural Networks (FBPNNs). Motivated by this inspiration, this paper proposes a state-of-the-art application of fractional calculus to implement a FBPNN trained by an improved FSDM whose reverse incremental search is in the negative directions of the approximate fractional-order partial derivatives of the square error. At first, the theoretical concept of a FBPNN trained by an improved FSDM is described mathematically. Then, the mathematical proof of the fractional-order global optimal convergence, an assumption of the structure, and the fractional-order multi-scale global optimization of a FBPNN trained by an improved FSDM are analysed in detail. Finally, we perform comparative experiments and compare a FBPNN trained by an improved FSDM with a classic first-order BPNN, i.e., an example function approximation, fractional-order multi-scale global optimization, and two comparative performances with real data. The more efficient optimal searching capability of the fractional-order multi-scale global optimization of a FBPNN trained by an improved FSDM to determine the global optimal solution is the major advantage being superior to a classic first-order BPNN.

## 1. Introduction

BACKPROPAGATION, an abbreviation for "backward propagation of errors", is a common method for training artificial neural networks and is used in conjunction with an optimization method such as gradient descent [1]. The classic first-order Backpropagation Algorithm (BA) to train multilayer networks was first described in the thesis of P. J. Werbos in 1974; the algorithm was presented in the context of general networks, with neural networks as a special case [2]. It was not until the mid-1980s that the classic BA was widely publicized. It was rediscovered independently by D. E. Rumelhart *et al.* [3], D. B. Parker [4], and Y. L. Cun [5]. The classic BA was popularized by its inclusion in the book Parallel Distributed Processing [6]. The publication of this book triggered a significant amount of research into neural networks. Multilayer perceptrons, trained by the classic BA, are currently the most widely used neural networks. The classic BA can refer to the result of a playout that is propagated up the search tree in a Monte Carlo tree search [7]. It has been demonstrated that classic two-layer first-order BackPropagation Neural Networks (BPNNs), with sigmoid activation functions in the hidden layer and linear transfer functions in the output layer, can approximate any function of interest to any degree of accuracy when a sufficient number of hidden units are available [8]. Research related to faster BPNN algorithms can be broadly classified into two categories. The first category involves the development of heuristic techniques. These heuristic techniques include concepts such as varying the learning rate using momentum and resealing variables [9]–[12]. The second category focuses on standard numerical optimization techniques. These techniques include concepts such as the conjugate gradient algorithm and Levenberg-Marquardt algorithm [13]–[17]. Combinations of neural networks and evolutionary computation procedures have been widely explored. This area addresses a wide range of topics such as cutting angle method [18], simulated annealing [19]–[21], swarm algorithms [22], genetic algorithms [23]–[26], and hybrid training methods [27]–[28], which are superior to traditional local techniques. In all these methods, the computational complexity increases rapidly with an increase in the number of variables. Moreover, the classic first-order BA-based BPNNs are easily trapped into a local optimal solution, whose optimization performance must be improved.

The application of fractional calculus to neural networks and cybernetics is an emerging discipline of research and a small number of studies have been conducted in this area. Fractional calculus has become an important novel branch in mathematical analyses [29], [30]. Recently, fractional calculus has become a promising mathematical method for physical scientists and engineering technicians. Promising results and ideas have demonstrated that fractional calculus can be an interesting and useful tool in many scientific fields such as diffusion processes [31], viscoelasticity theory [32], fractal dynamics [33], fractional control [34], [35], fractance [36], fracmemristor [37], image processing [38], [39], and neural networks [40]–[43]. Fractional calculus has been applied to neural networks and cybernetics primarily owing to its inherent advantages of long-term memory, nonlocality, and weak singularity, important properties of fractional calculus [29], [30]. The basic characteristic feature of fractional calculus is that it extends the concepts of the integer-order difference and Riemann sums. The characteristics of fractional calculus are considerably different from those of classic integer-order calculus. For example the fractional differential, except based on the Caputo definition, of a Heaviside function is nonzero, whereas its integer-order differential must be zero [29], [30]. Thus, the properties of the Fractional-order Steepest Descent Method (FSDM) are also different from those of the traditional first-order steepest descent method [40]. For example, the

Yi-Fei PU is a professor and doctoral supervisor with College of Computer Science, Sichuan University, Chengdu 610065, China (corresponding author, e-mail: puyifei@scu.edu.cn).

Jian Wang is an associate professor and doctoral supervisor with School of Science, China University of Petroleum, Qingdao, 266580, China (corresponding author, e-mail: wangjiannl@upc.edu.cn). He is an associate editor of IEEE Transactions on Neural Networks and Learning Systems.

FSDM can determine the fractional-order extreme points of the energy norm, which do not overlap the traditional first-order stationary points [40]. It is known that classic first-order BPNNs demonstrate a tendency to be trapped into a local optimal solution.

The application of the improved FSDM to training BPNNs has the potential of overcoming such deficits. Therefore, to improve the optimization performance of classic first-order BPNNs, in this paper we study whether it could be possible to apply improved FSDM to generalize classic first-order BPNNs to the FBPNNs. Based on this inspiration, in this study, we introduce a FBPNN trained by an improved FSDM, whose reverse incremental search is in the negative directions of the approximate fractional-order partial derivatives of the square error at iteration $k$ of the iterative search process. This introduced fractional-order branch of the family of BPNNs trained by the improved FSDM differs from the majority of the previous classic first-order BPNNs and as such trained by the traditional first-order steepest descent method, which represents an interesting theoretical contribution. The more efficient optimal searching capability of the fractional-order multi-scale global optimization of a FBPNN trained by an improved FSDM to determine the global optimal solution is the major advantage being superior to a classic first-order BPNN.

The remainder of this manuscript is organized as follows: In Section 2, the necessary theoretical background for fractional calculus and the FSDM is presented. In Section 3, the theoretical concept of a FBPNN trained by an improved FSDM is discussed in detail. At first, we describe the fractional-order partial derivatives of the square error of a BPNN. Secondly, the achievement of the improved FSDM for the family of the BPNNs and the improved FSDM based FBPNNs is further discussed. Thirdly, the mathematical proof of the fractional-order global optimal convergence, an assumption of the structure, and the fractional-order multi-scale global optimization of a FBPNN trained by an improved FSDM are analysed. The experimental results and analysis are reported in Section 4. Section 4 presents a sample function approximation, a fractional-order multi-scale global optimization, and two comparative performances with real data. In Section 5, the conclusions of this manuscript are presented.

## 2. Related Work

This section includes a brief introduction to the necessary mathematical background of fractional calculus and the FSDM.

The commonly used fractional calculus definitions are Grünwald-Letnikov, Riemann-Liouville, and Caputo [29], [30]. The Grünwald-Letnikov definition of fractional calculus for causal signal $f(x)$ can be represented in a convenient form as follows:

$$^{G-L}_{\ a}D^v_x f(x) = \lim_{N\to\infty}\left\{\frac{((x-a)/N)^{-v}}{\Gamma(-v)}\sum_{k=0}^{N-1}\frac{\Gamma(k-v)}{\Gamma(k+1)}f\left(x-k\left(\frac{x-a}{N}\right)\right)\right\}, \quad (1)$$

where $f(x)$ is a differintegrable function [29], [30], $[a,x]$ is the duration of $f(x)$, $N$ is the number of partitions of the duration,

$v$ is an arbitrary real number, $\Gamma(\alpha)=\int_0^\infty e^{-x}x^{\alpha-1}dx$ is gamma function and $^{G-L}_{\ a}D^v_x$ denotes the Grünwald-Letnikov defined fractional differential operator. In this manuscript, we use the equivalent notations $D^v_x = ^{G-L}_{\ a}D^v_x$ in an interchangeable manner.

Secondly, the reverse incremental search of the FSDM is in the negative direction of the $v$-order fractional derivative of the quadratic energy norm $E$, which can be represented as [40]:

$$\xi_{k+1} = \xi_k - \mu\left(D^v_{\xi_k}E\right), \quad (2)$$

where $k$ is the step size or number of iterations, $\xi$ is the independent variable of $E$, $D^v_{\xi_k}$ is the $v$-order fractional derivative of $E$ at $\xi = \xi_k$, and $\mu$ is the constant coefficient that controls the stability and the rate of convergence of the FSDM.

## 3. Fractional-order Backpropagation Neural Networks

### 3.1 Fractional-order partial derivatives of square error

In this subsection, to achieve the improved FSDM for training the FBPNNs, the fractional-order partial derivatives of the square error of a BPNN should be described firstly.

Figure 1 displays the model of a BPNN, which is represented by abbreviated symbols denoting its three layers.

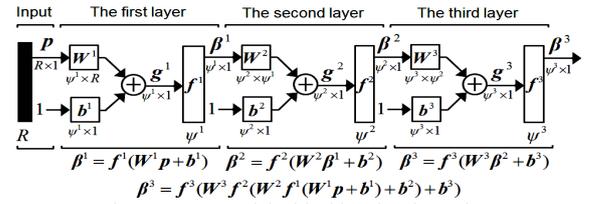

$$\beta^1 = f^1(W^1\beta^0 + b^1)$$
$$\beta^2 = f^2(W^2\beta^1 + b^2) \quad \beta^3 = f^3(W^3\beta^2 + b^3)$$
$$\beta^3 = f^3(W^3f^2(W^2f^1(W^1\beta^0 + b^1) + b^2) + b^3)$$

Fig. 1. BPNN model with abbreviated notation.

In Fig. 1, the superscript denotes the number of layers of a BPNN. $p_{R\times 1}$ is the input matrix of a BPNN. $W^1_{\psi^1\times R}$, $W^2_{\psi^2\times\psi^1}$, and $W^3_{\psi^3\times\psi^2}$ are the weight matrices of its first, second, and third layers, respectively. $b^1_{\psi^1\times 1}$, $b^2_{\psi^2\times 1}$, and $b^3_{\psi^3\times 1}$ are the bias matrices of its first, second, and third layers, respectively. $f^1$, $f^2$, and $f^3$ are the activation functions of its first, second, and third layer, respectively. $g^1_{\psi^1\times 1}$, $g^2_{\psi^2\times 1}$, and $g^3_{\psi^3\times 1}$ are the net input matrices of $f^1$, $f^2$, and $f^3$, respectively. $\beta^1_{\psi^1\times 1}$, $\beta^2_{\psi^2\times 1}$, and $\beta^3_{\psi^3\times 1}$ are the output matrices of the first, second, and third layers of a BPNN, respectively. In a multi-layer BPNN, the output of its upper layer is the input of the next layer. Thus, it follows that:

$$\beta^{m+1} = f^{m+1}(g^{m+1}) = f^{m+1}(W^{m+1}\beta^m + b^{m+1}), \quad m = 0,1,\cdots,M-1, \quad (3)$$

where $M$ is the number of layers of a BPNN and $g^{m+1} = W^{m+1}\beta^m + b^{m+1}$, $\beta^0 = p_{R\times 1}$, and $\beta = \beta^M$ represent the actual output of the entire BPNN. Assume that the sample sets of the target output matrix $q_\eta$ corresponding to the actual input matrix $p_\eta$ of a BPNN are $\{p_1,q_1\},\{p_2,q_2\},\cdots,\{p_\eta,q_\eta\}$. Then, the mean square error of a BPNN is as follows:

$$F(x) = E(e^T e) = E[(q-\beta)^T(q-\beta)],$$ (4)

where $x$ is the weight and bias vector and $e = [e_i]$ is the error vector of a BPNN. A BPNN uses the iterative approximate calculation of variance as follows:

$$\hat{F}(k) = [q(k) - \beta(k)]^T[q(k) - \beta(k)] = e(k)^T e(k),$$ (5)

where $k$ is the number of iterations, and $\hat{F}(k) = \sum_{j=1}^{w^s}(q_j - \beta_j)^2$.

Equation (5) indicates that the mathematical expectation of the mean squared error $F$ is replaced by a quadratic energy norm $\hat{F}(k)$, the square error at iteration $k$ of the iterative search process, of a BPNN.

In addition, $\sum_{m=0}^{\infty}\sum_{u=0}^{\infty} = \sum_{u=0}^{\infty}\sum_{m=0}^{\infty}$ and $\binom{v}{r+n}\binom{r+n}{n} = \binom{v}{n}\binom{v-n}{r}$ can be derived, where $v$ is a fraction and $\binom{v}{n} = \frac{(-1)^n\Gamma(n-v)}{\Gamma(-v)\Gamma(1+n)} = \frac{\Gamma(1+v)}{\Gamma(1-n+v)\Gamma(1+n)}$. Thus, from (1), it follows that:

$$D_{x-a}^v(\hat{F}g_i^m) = \sum_{n=0}^{\infty}\left[\binom{v}{n}\left(D_{x-a}^{v-n}\hat{F}\right)\left(D_{x-a}^n g_i^m\right)\right],$$ (6)

where $x$ is an independent variable, $g_i^m$ is the $i^{th}$ net input of $f^m$, and $a$ is a constant. When $n > v$, $\binom{v}{n} = \frac{(-1)^n\Gamma(n-v)}{\Gamma(-v)\Gamma(n+1)} = \frac{\Gamma(1+v)}{\Gamma(1-n+v)\Gamma(n+1)} \neq 0$ [29]. Thus, from (6) and Faà di Bruno's formula, the fractional-order partial derivatives of the square error of a BPNN can be derived as:

$$D_{(w_{i,j}^m - w_{int})}^v\hat{F}[g_i^m(w_{i,j}^m, b_i^m)] = D_{g_i^m}^v\hat{F}[g_i^m(w_{i,j}^m, b_i^m)]$$
$$= \frac{(w_{i,j}^m - w_{int})^{-v}}{\Gamma(1-v)}\hat{F} + \sum_{m=1}^{\infty}\sum_{n=0}^{\infty}\binom{v}{n}\frac{(w_{i,j}^m - w_{int})^{n-v}}{\Gamma(n-v+1)}n!\left(D_{g_i^m}^n\hat{F}\right)\sum\prod_{k=1}^{n}\frac{1}{P_k!}\left[\frac{D_{w_{i,j}^m}^k g_i^m}{k!}\right]^{P_k},$$ (7)

$$D_{(b_i^m - b_{int})}^v\hat{F}[g_i^m(w_{i,j}^m, b_i^m)] = D_{g_i^m}^v\hat{F}[g_i^m(w_{i,j}^m, b_i^m)]$$
$$= \frac{(b_i^m - b_{int})^{-v}}{\Gamma(1-v)}\hat{F} + \sum_{m=1}^{\infty}\sum_{n=0}^{\infty}\binom{v}{n}\frac{(b_i^m - b_{int})^{n-v}}{\Gamma(n-v+1)}n!\left(D_{g_i^m}^n\hat{F}\right)\sum\prod_{k=1}^{n}\frac{1}{P_k!}\left[\frac{D_{b_i^m}^k g_i^m}{k!}\right]^{P_k},$$ (8)

where $w_{i,j}^m \in (w_{int}, w_{sup})$ and $b_i^m \in (b_{int}, b_{sup})$ are the domain of definition of $w_{i,j}^m$ and $b_i^m$, respectively. $\hat{F} = \hat{F}[g_i^m(w_{i,j}^m, b_i^m)]$ is a composite function, $D_{g_i^m}^h$, $D_{w_{i,j}^m}^k$, and $D_{b_i^m}^k$ are the integer-order differential operators, and nonnegative integer $P_k$ satisfies:

$$\begin{cases} \sum_{k=1}^{n} kP_k = n \\ \sum_{k=1}^{n} P_k = h \end{cases}.$$ (9)

The third summation notations $\sum$ in (7) and (8) denote the summation of the corresponding $\left\{\prod_{k=1}^{n}\frac{1}{P_k!}\left[\left(D_{w_{i,j}^m}^k g_i^m\right)/k!\right]^{P_k}\right\}\Big|_{h=1\rightarrow\infty}$ and $\left\{\prod_{k=1}^{n}\frac{1}{P_k!}\left[\left(D_{b_i^m}^k g_i^m\right)/k!\right]^{P_k}\right\}\Big|_{h=1\rightarrow\infty}$ of all of the combinations of $P_k|_{k=1\rightarrow n}$ that satisfy the requirement of (9).

From Fig. 1, we can observe that $g^{m+1} = W^{m+1}\beta^m + b^{m+1}$. Thus, the net input $g_i^m$ of $f^m$ is as follows:

$$g_i^m(w_{i,j}^m, b_i^m) = \sum_{j=1}^{w^{m-1}} w_{i,j}^m\beta_j^{m-1} + b_i^m.$$ (10)

From (10), it follows that:

$$D_{w_{i,j}^m}^1 g_i^m(w_{i,j}^m, b_i^m) = \beta_j^{m-1},$$ (11)

$$D_{b_i^m}^1 g_i^m(w_{i,j}^m, b_i^m) = 1.$$ (12)

From (11) and (12), it follows that:

$$D_{w_{i,j}^m}^{k>1} g_i^m(w_{i,j}^m, b_i^m) \equiv 0,$$ (13)

$$D_{b_i^m}^{k>1} g_i^m(w_{i,j}^m, b_i^m) \equiv 0.$$ (14)

Using mathematical induction, it can be proven that from (9), we can derive that $\begin{cases} P_{1\leq i\leq n-1} = 0 \\ P_n = 1 \end{cases}$, when $h=1$, $\begin{cases} P_{1\leq i\leq n+1,\, i\neq j-1,\, n-j} = 0 \\ P_{1+j-1} = P_{n-j} = 1 \end{cases}$, $j=1\cdots n-1$, when $h=2$, $\cdots$, and $\begin{cases} P_2 = P_3 = \cdots = P_n = 0 \\ P_1 = n \end{cases}$, when $h=n$. Thus, from (7) and (8), if and only if, $D_{w_{i,j}^m}^1 g_i^m(w_{i,j}^m, b_i^m) \neq 0$, $D_{b_i^m}^1 g_i^m(w_{i,j}^m, b_i^m) \neq 0$, $D_{w_{i,j}^m}^{k>1} g_i^m(w_{i,j}^m, b_i^m) \equiv 0$, and $D_{b_i^m}^{k>1} g_i^m(w_{i,j}^m, b_i^m) \equiv 0$, only when $h=n$, we obtain $\sum\prod_{k=1}^{n}\left[\frac{D_{w_{i,j}^m}^k g_i^m}{k!}\right]^{P_k} \neq 0$ and $\sum\prod_{k=1}^{n}\frac{1}{P_k!}\left[\frac{D_{b_i^m}^k g_i^m}{k!}\right]^{P_k} \neq 0$. Thus, from (11)–(14), (7) and (8) can be simplified as follows:

$$D_{w_{i,j}^m}^v\hat{F}[g_i^m(w_{i,j}^m, b_i^m)] = \frac{(w_{i,j}^m - w_{int})^{-v}}{\Gamma(1-v)}\hat{F} + \sum_{m=1}^{\infty}\binom{v}{n}\frac{(w_{i,j}^m - w_{int})^{n-v}}{\Gamma(n-v+1)}n!\left(D_{g_i^m}^n\hat{F}\right)\frac{1}{P_1!}\left(D_{w_{i,j}^m}^1 g_i^m\right)^{P_1}\Big|_{h=n},$$ (15)

$$D_{b_i^m}^v\hat{F}[g_i^m(w_{i,j}^m, b_i^m)] = \frac{(b_i^m - b_{int})^{-v}}{\Gamma(1-v)}\hat{F} + \sum_{m=1}^{\infty}\binom{v}{n}\frac{(b_i^m - b_{int})^{n-v}}{\Gamma(n-v+1)}n!\left(D_{g_i^m}^n\hat{F}\right)\frac{1}{P_1!}\left(D_{b_i^m}^1 g_i^m\right)^{P_1}\Big|_{h=n}.$$ (16)

From (9), when $h=n$, it follows that:

$$\begin{cases} P_1 = n \\ P_2 = P_3 = \cdots = P_{n-1} = P_n = 0 \end{cases}$$ (17)

Thus, from (11), (12), and (17), (15) and (16) can be further simplified as follows:

$$D_{w_{i,j}^m}^v\hat{F}[g_i^m(w_{i,j}^m, b_i^m)] = \frac{(w_{i,j}^m - w_{int})^{-v}}{\Gamma(1-v)}\hat{F} + \sum_{n=1}^{\infty}\binom{v}{n}\frac{(w_{i,j}^m - w_{int})^{n-v}}{\Gamma(n-v+1)}\left(D_{g_i^m}^n\hat{F}\right)\left(D_{w_{i,j}^m}^1 g_i^m\right)^n$$
$$= \frac{(w_{i,j}^m - w_{int})^{-v}}{\Gamma(1-v)}\hat{F} + \sum_{n=1}^{\infty}\binom{v}{n}\frac{(w_{i,j}^m - w_{int})^{n-v}}{\Gamma(n-v+1)}\left(D_{g_i^m}^n\hat{F}\right)(\beta_j^{m-1})^n,$$ (18)

$$D_{b_i^m}^v\hat{F}[g_i^m(w_{i,j}^m, b_i^m)] = \frac{(b_i^m - b_{int})^{-v}}{\Gamma(1-v)}\hat{F} + \sum_{n=1}^{\infty}\binom{v}{n}\frac{(b_i^m - b_{int})^{n-v}}{\Gamma(n-v+1)}\left(D_{g_i^m}^n\hat{F}\right)\left(D_{b_i^m}^1 g_i^m\right)^n$$
$$= \frac{(b_i^m - b_{int})^{-v}}{\Gamma(1-v)}\hat{F} + \sum_{n=1}^{\infty}\binom{v}{n}\frac{(b_i^m - b_{int})^{-v}}{\Gamma(n-v+1)}\left(D_{g_i^m}^n\hat{F}\right).$$ (19)

Note that at first, (18) and (19) are the chain rule of the fractional derivatives of the square error (a composite function) of a BPNN [29]. From (18) and (19), we can observe that the chain rule of the fractional derivatives of the square error of a BPNN gives an infinite series that offers minimal expectation of being expressible in closed form, except for trivially simple instances of the functions $\hat{F}$ and $g_i^m$. When (18) and (19) are a zero initial condition, they continue to be set up and practical [29]. Secondly, the issue of the nonzero initial value problem is a fundamental issue of the application of fractional calculus. The practical applicability of (18) and (19) is limited by the absence of the physical interpretation of the limit values of fractional derivatives at the lower bound $w_{i,j}^m = w_{int}$ and $b_i^m = b_{int}$, respectively. To date, such interpretation was partially solved in the work of Heymans and Podlubny [44], [45]. Finally, the summation notations $\sum\limits_{n=1}^{\infty}$ in (18) and (19) denote that both $D_{w_{i,j}^m}^{v}\hat{F}\left[g_i^m\left(w_{i,j}^m,b_i^m\right)\right]$ and $D_{b_i^m}^{v}\hat{F}\left[g_i^m\left(w_{i,j}^m,b_i^m\right)\right]$ express the long-term memory and nonlocality of the fractional differential of the square error $\hat{F}$ of a BPNN.

### 3.2 Achievement of improved FSDM and FBPNNs trained by improved FSDM

In this subsection, the achievement of the improved FSDM and the FBPNNs trained by the improved FSDM is further discussed.

To simplify the infinite series calculation of the fractional-order partial derivative of the square error of a BPNN in (18) and (19), the approximate fractional-order partial derivatives of the square error of a BPNN, $\tilde{D}_{w_{i,j}^m}^{v}\hat{F}\left[g_i^m\left(w_{i,j}^m,b_i^m\right)\right]$ and $\tilde{D}_{b_i^m}^{v}\hat{F}\left[g_i^m\left(w_{i,j}^m,b_i^m\right)\right]$, are suggested to merely take the first four terms of (18) and (19) into account, which can be given as:

$$\tilde{D}_{w_{i,j}^m}^{v}\hat{F}\left[g_i^m\left(w_{i,j}^m,b_i^m\right)\right]=\frac{\left(w_{i,j}^m-w_{\text{inf}}\right)^{-v}}{\Gamma(1-v)}\hat{F}+\sum_{n=1}^{3}\binom{v}{n}\frac{\left(w_{i,j}^m-w_{\text{inf}}\right)^{n-v}}{\Gamma(n-v+1)}\left(D_{g_i^m}^{n}\hat{F}\right)\left(\beta_j^{m-1}\right)^n,\quad(20)$$

$$\tilde{D}_{b_i^m}^{v}\hat{F}\left[g_i^m\left(w_{i,j}^m,b_i^m\right)\right]=\frac{\left(b_i^m-b_{\text{inf}}\right)^{-v}}{\Gamma(1-v)}\hat{F}+\sum_{n=1}^{3}\binom{v}{n}\frac{\left(b_i^m-b_{\text{inf}}\right)^{n-v}}{\Gamma(n-v+1)}\left(D_{g_i^m}^{n}\hat{F}\right).\quad(21)$$

Equations (20) and (21) indicate that although the first four terms of $D_{w_{i,j}^m}^{v}\hat{F}\left[g_i^m\left(w_{i,j}^m,b_i^m\right)\right]$ and $D_{b_i^m}^{v}\hat{F}\left[g_i^m\left(w_{i,j}^m,b_i^m\right)\right]$ are only take into account, $\tilde{D}_{w_{i,j}^m}^{v}\hat{F}\left[g_i^m\left(w_{i,j}^m,b_i^m\right)\right]$ and $\tilde{D}_{b_i^m}^{v}\hat{F}\left[g_i^m\left(w_{i,j}^m,b_i^m\right)\right]$ are essentially the approximate fractional differentials of the square error $\hat{F}$ of a BPNN, which still preserve the strengths of fractional calculus such as long-term memory, nonlocality, and weak singularity. Therefore, in a similar way to the FSDM, from (2), (20), and (21), the improved FSDM for the family of the BPNNs can be implemented as follows:

$$w_{i,j}^m(k+1)=w_{i,j}^m(k)-\mu\left[\tilde{D}_{w_{i,j}^m}^{v}\hat{F}(k)\right],\quad(22)$$

$$b_i^m(k+1)=b_i^m(k)-\mu\left[\tilde{D}_{b_i^m}^{v}\hat{F}(k)\right],\quad(23)$$

where $w_{i,j}^m$ is the weight between the $j^{th}$ output of the $(m-1)^{th}$ layer and the $i^{th}$ input of the $m^{th}$ layer, $b_i^m$ is the $i^{th}$ bias of the $m^{th}$ layer, and $\mu$ is the learning rate of an improved FSDM (a positive small number). Note that in (22) and (23), even if $\tilde{D}_{w_{i,j}^m}^{v}\hat{F}(k)=0$ and $\tilde{D}_{b_i^m}^{v}\hat{F}(k)=0$, but $\hat{F}(k)\neq0$, the iterative search processes should be constrained to keep going on. When $\tilde{D}_{w_{i,j}^m}^{v}\hat{F}(k)=0$ and $\tilde{D}_{b_i^m}^{v}\hat{F}(k)=0$, but $\hat{F}(k)\neq0$, the corresponding point $\left(w_{i,j}^{m*},b_i^{m*}\right)$ is merely a saddle point, rather than a real fractional-order optimal minimum point of an improved FSDM. Equations (22) and (23) can be expressed in vector form, $W^m(k+1)=W^m(k)-\mu\tilde{D}_{w^m}^{v}\hat{F}$ and $b^m(k+1)=b^m(k)-\mu\tilde{D}_{b^m}^{v}\hat{F}$, where $\tilde{D}_{w^m}^{v}\hat{F}=\left[\tilde{D}_{w_{i,j}^m}^{v}\hat{F}\right]_{p^m\times p^{m-1}}$ and $\tilde{D}_{b^m}^{v}\hat{F}=\left[\tilde{D}_{b_i^m}^{v}\hat{F}\right]_{p^m\times 1}$. We define the $n$-order sensibility ${}_{n}\rho_i^m(k)$ at iteration $k$ of the iterative search

process of the improved FSDM for the family of the BPNNs as follows:

$${}_{n}\rho_i^m(k)=D_{g_i^m}^{n}\hat{F}(k).\quad(24)$$

Thus, from (24), (20) and (21) can be rewritten as follows:

$$\tilde{D}_{w_{i,j}^m}^{v}\hat{F}(k)=\frac{\left[w_{i,j}^m(k)-w_{\text{inf}}\right]^{-v}}{\Gamma(1-v)}\hat{F}(k)+\sum_{n=1}^{3}\binom{v}{n}\frac{\left[w_{i,j}^m(k)-w_{\text{inf}}\right]^{n-v}}{\Gamma(n-v+1)}{}_{n}\rho_i^m(k)\left(\beta_j^{m-1}\right)^n,\quad(25)$$

$$\tilde{D}_{b_i^m}^{v}\hat{F}(k)=\frac{\left[b_i^m(k)-b_{\text{inf}}\right]^{-v}}{\Gamma(1-v)}\hat{F}(k)+\sum_{n=1}^{3}\binom{v}{n}\frac{\left[b_i^m(k)-b_{\text{inf}}\right]^{n-v}}{\Gamma(n-v+1)}{}_{n}\rho_i^m(k).\quad(26)$$

Thus, from (22), (23), (25), and (26), it follows that:

$$w_{i,j}^m(k+1)=w_{i,j}^m(k)-\mu\left\{\frac{\left[w_{i,j}^m(k)-w_{\text{inf}}\right]^{-v}}{\Gamma(1-v)}\hat{F}(k)+\sum_{n=1}^{3}\binom{v}{n}\frac{\left[w_{i,j}^m(k)-w_{\text{inf}}\right]^{n-v}}{\Gamma(n-v+1)}{}_{n}\rho_i^m(k)\left(\beta_j^{m-1}\right)^n\right\},\quad(27)$$

$$b_i^m(k+1)=b_i^m(k)-\mu\left\{\frac{\left[b_i^m(k)-b_{\text{inf}}\right]^{-v}}{\Gamma(1-v)}\hat{F}(k)+\sum_{n=1}^{3}\binom{v}{n}\frac{\left[b_i^m(k)-b_{\text{inf}}\right]^{n-v}}{\Gamma(n-v+1)}{}_{n}\rho_i^m(k)\right\}.\quad(28)$$

Note that (27) and (28), and Fig. 1, indicate that by employing the improved FSDM to achieve the training process for the family of the BPNNs, a BPNN is indeed a FBPNN trained by an improved FSDM. The architecture of a FBPNN trained by an improved FSDM is identical to that of a traditional first-order BPNN; however, the reverse incremental searches of a FBPNN trained by an improved FSDM are in the negative directions of the approximate $v$-order fractional derivatives of quadratic energy norm $\hat{F}(k)$. As we know, the classic first-order BA for the approximate mean squared error of a first-order BPNN is as follows [1]–[6]:

$$w_{i,j}^m(k+1)=w_{i,j}^m(k)-\mu\left[D_{w_{i,j}^m}^{1}\hat{F}(k)\right]=w_{i,j}^m(k)-\mu_1\rho_i^m(k)\beta_j^{m-1},\quad(29)$$

$$b_i^m(k+1)=b_i^m(k)-\mu\left[D_{b_i^m}^{1}\hat{F}(k)\right]=b_i^m(k)-\mu_1\rho_i^m(k),\quad(30)$$

where $D_{w_{i,j}^m}^{1}\hat{F}(k)={}_{1}\rho_i^m(k)\beta_j^{m-1}=\left[D_{g_i^m}^{1}\hat{F}(k)\right]\beta_j^{m-1}$ and ${}_{1}\rho_i^m(k)=D_{g_i^m}^{1}\hat{F}(k)$. The reverse incremental search of a classic first-order BPNN is in the negative direction of the first-order derivative of the square error $\hat{F}(k)$. Comparing (27) and (28) with (29) and (30), we can observe that when $v=1$, a FBPNN trained by an improved FSDM converts to be a classic first-order BPNN. The first-order BPNN is a special case of a FBPNN trained by an improved FSDM. The classic first-order optimal minimum point is a particular case of the fractional-order minimum one [40].

### 3.3 Fractional-order global optimal convergence of improved FSDM based FBPNN

In this subsection, the mathematical proof of the fractional-order global optimal convergence of a FBPNN trained by an improved FSDM is presented.

Assume that the square error $\hat{F}(k)$ of a FBPNN trained by an improved FSDM is a smooth convex function with at least one of the equivalent fractional-order optimal extreme points $\left(w_{i,j}^{m*},b_i^{m*}\right)$. Each iterative step of the iterative search processes of a FBPNN trained by an improved FSDM is formulated as (22), (23), (27) and (28). The fingerprint of a fractional-order optimal minimum point is its $\tilde{D}_{w_{i,j}^m}^{v}\hat{F}(k)$ and $\tilde{D}_{b_i^m}^{v}\hat{F}(k)$ are equal to zero. Therefore, we can obtain the following lemma.

***Lemma 1***: If the number of neurons and number of hidden layers of a FBPNN trained by an improved FSDM are such that at least one fractional-order minimum point of the square error $\hat{F}(k)$ exists; even if $\tilde{D}_{w_{i,j}^m}^\nu \hat{F}(k) = 0$ and $\tilde{D}_{b_i^m}^\nu \hat{F}(k) = 0$, but $\hat{F}(k) \neq 0$, the iterative search processes are constrained to keep going on, then the improved FSDM described in (22), (23), (27) and (28) can guarantee this FBPNN converge to a fractional-order optimal minimum point $\left(w_{i,j}^{m*}, b_i^{m*}\right)$, a global minimum point, of the square error $\hat{F}(k)$.

***Proof***: Equations (22), (23), and (25)–(28) indicate that on a fractional-order optimal minimum point $\left(w_{i,j}^{m*}, b_i^{m*}\right)$ of the square error $\hat{F}(k)$, one can obtain $\tilde{D}_{w_{i,j}^m}^\nu \hat{F}(k) = 0$ and $\tilde{D}_{b_i^m}^\nu \hat{F}(k) = 0$. Thus, the criteria for the convergence of (27) and (28) are:

$$\begin{cases} \tilde{D}_{w_{i,j}^m}^\nu \hat{F}(k) = \left\{ \dfrac{[w_{i,j}^m(k) - w_{\inf}]^{-\nu}}{\Gamma(1-\nu)} \hat{F}(k) + \sum_{n=1}^3 \binom{\nu}{n} \dfrac{[w_{i,j}^m(k) - w_{\inf}]^{n-\nu}}{\Gamma(n-\nu+1)} {}_n\rho_i^m(k) \left(\beta_j^{m-1}\right)^n \right\} = 0 \\ \tilde{D}_{b_i^m}^\nu \hat{F}(k) = \dfrac{[b_i^m(k) - b_{\inf}]^{-\nu}}{\Gamma(1-\nu)} \hat{F}(k) + \sum_{n=1}^3 \binom{\nu}{n} \dfrac{[b_i^m(k) - b_{\inf}]^{n-\nu}}{\Gamma(n-\nu+1)} {}_n\rho_i^m(k) \end{cases} \quad (31)$$

Equation (31) is the criterion for the convergence of (22), (23), (27) and (28). Note that as aforementioned discussion, in (22) and (23), even if $\tilde{D}_{w_{i,j}^m}^\nu \hat{F}(k) = 0$ and $\tilde{D}_{b_i^m}^\nu \hat{F}(k) = 0$, but $\hat{F}(k) \neq 0$, the iterative search processes are constrained to keep going on. Therefore, when $\tilde{D}_{w_{i,j}^m}^\nu \hat{F}(k) = 0$ and $\tilde{D}_{b_i^m}^\nu \hat{F}(k) = 0$, but $\hat{F}(k) \neq 0$, the corresponding point $\left(w_{i,j}^{m*}, b_i^{m*}\right)$ is merely a saddle point, rather than a real fractional-order optimal minimum point $\left(w_{i,j}^{m*}, b_i^{m*}\right)$ of an improved FSDM. The solution of (31) depends on both the square error $\hat{F}(k)$ and the $n$-order sensibility ${}_n\rho_i^m(k)$ at iteration $k$ of the iterative search process of a FBPNN trained by an improved FSDM.

For a FBPNN trained by an improved FSDM, because the bias $\boldsymbol{b}_{w^{m+1} \times 1}^{m+1}$ of the $(m+1)th$ layer is not related to the net input $\boldsymbol{g}_{w^{m+1}}^{m+1}$ of the $m th$ layer, from (3) and (10), we can derive that:

$$D_{g_j^m}^1 g_i^{m+1} = w_{i,j}^{m+1}\left(D_{g_j^m}^1 \beta_j^m\right) = w_{i,j}^{m+1}\left[D_{g_j^m}^1 f^m(g_j^m)\right] = w_{i,j}^{m+1}\left[D_{g_j^m}^1 f^m\left(\sum_{k=1}^{w^{m-1}} w_{j,k}^m \beta_k^{m-1} + b_j^m\right)\right] \quad (32)$$

Thus, from (32), it follows that:

$$\begin{aligned} {}_2\rho_i^m(k) &= \left(D_{g_i^m}^1 g_i^{m+1}\right)^T \left[D_{g_i^{m+1}}^1\left(D_{g_i^m}^1 \hat{F}(k)\right)\right] = \left(D_{g_i^m}^1 g_i^{m+1}\right)^T \left[D_{g_i^m}^1\left(\left(D_{g_i^m}^1 g_i^{m+1}\right)\left(D_{g_i^{m+1}}^1 \hat{F}(k)\right)\right)\right] \quad (33) \\ &= \left(D_{g_i^m}^1 g_i^{m+1}\right)^2 \left(D_{g_i^{m+1}}^2 \hat{F}(k)\right) = \left\{ w_{i,j}^{m+1}\left[D_{g_i^m}^1 f^m\left(\sum_{k=1}^{w^{m-1}} w_{j,k}^m \beta_k^{m-1} + b_j^m\right)\right] \right\}^2 {}_2\rho_i^{m+1}(k). \end{aligned}$$

which is closely related to the activation function of the $m^{th}$ layer of a FBPNN trained by an improved FSDM, $f^m$. In a similar to (33), from (24) and Fig.1, we can derive that:

$${}_n\rho_i^m(k) = \left(D_{g_i^m}^1 g_i^{m+1}\right)^n \left(D_{g_i^{m+1}}^n \hat{F}(k)\right) = \left\{ w_{i,j}^{m+1}\left[D_{g_i^m}^1 f^m\left(\sum_{k=1}^{w^{m-1}} w_{j,k}^m \beta_k^{m-1} + b_j^m\right)\right] \right\}^n {}_n\rho_i^{m+1}(k). \quad (34)$$

Equation (34) indicates that the sensibility back propagates from the last layer to the first layer of the improved FSDM based FBPNN. In particular, with regard to the $M^{th}$ layer, the last layer, of a FBPNN, from (3), (5), and (24), we can derive that:

$${}_1\rho_i^M(k) = D_{g_i^M}^1 \hat{F}(k) = D_{g_i^M}^1 \left[\sum_{j=1}^{w^M}(q_j - \beta_j)^2\right] = -2(q_i - \beta_i)\left(D_{g_i^M}^1 \beta_i\right), \quad (35)$$

$$D_{g_i^M}^1 \beta_i = D_{g_i^M}^1 \beta_i^M = D_{g_i^M}^1 f^M\left(g_i^M\right). \quad (36)$$

Substitute (36) into (35), we obtain:

$${}_1\rho_i^M(k) = -2(q_i - \beta_i)\left(D_{g_i^M}^1 \beta_i\right) = -2(q_i - \beta_i)\left[D_{g_i^M}^1 f^M\left(g_i^M\right)\right] \quad (37)$$

From (24), (35), and (37), we have:

$${}_2\rho_i^M(k) = D_{g_i^M}^1\left[{}_1\rho_i^M(k)\right] = -2\left[(q_i - \beta_i)\left(D_{g_i^M}^2 \beta_i\right) - \left(D_{g_i^M}^1 \beta_i\right)^2\right], \quad (38)$$

$${}_3\rho_i^M(k) = D_{g_i^M}^1\left[{}_2\rho_i^M(k)\right] = -2\left[(q_i - \beta_i)\left(D_{g_i^M}^3 \beta_i\right) - 3\left(D_{g_i^M}^1 \beta_i\right)\left(D_{g_i^M}^2 \beta_i\right)\right]. \quad (39)$$

Equations (37)–(39) are the initial point values of the backpropagation recurrence relations, which can be expressed by (34).

In addition, if $\tilde{D}_{w_{i,j}^m}^\nu \hat{F}(k) = 0$ and $\tilde{D}_{b_i^m}^\nu \hat{F}(k) = 0$, but $\hat{F}(k) \neq 0$, as long as the iterative search processes are constrained to keep going on, then, when the iterative search processes converge to to a fractional-order optimal minimum point $\left(w_{i,j}^{m*}, b_i^{m*}\right)$ of the square error $\hat{F}(k)$, with respect to the $M^{th}$ layer (the last layer) of a FBPNN trained by an improved FSDM, to enable (31) to be set up, from (37), (38), and (39), a sufficient and necessary condition can be given as:

$$\begin{cases} \hat{F}(k) = \sum_{j=1}^M (q_j - \beta_j)^2 = 0 \\ {}_1\rho_i^M(k) = -2(q_i - \beta_i)\left(D_{g_i^M}^1 \beta_i\right) = 0 \\ {}_2\rho_i^M(k) = -2\left[(q_i - \beta_i)\left(D_{g_i^M}^2 \beta_i\right) - \left(D_{g_i^M}^1 \beta_i\right)^2\right] = 0 \\ {}_3\rho_i^M(k) = -2\left[(q_i - \beta_i)\left(D_{g_i^M}^3 \beta_i\right) - 3\left(D_{g_i^M}^1 \beta_i\right)\left(D_{g_i^M}^2 \beta_i\right)\right] \end{cases} \quad (40)$$

where $\beta_i = \beta_i^M$ and $i = 1, 2, \cdots, w^M$. To enable (40) to be set up, a sufficient and necessary condition can be given as:

$$\begin{cases} q_i - \beta_i = 0 \\ D_{g_i^M}^1 \beta_i = 0 \end{cases}. \quad (41)$$

Further, when (41) is set up, we have:

$$\begin{cases} \sum_{j=1}^M (q_j - \beta_j)^2 = 0 \\ -2(q_i - \beta_i)\left(D_{g_i^M}^1 \beta_i\right) = 0 \end{cases}. \quad (42)$$

Therefore, when the iterative search processes converge to a fractional-order optimal minimum point $\left(w_{i,j}^{m*}, b_i^{m*}\right)$ of the square error $\hat{F}(k)$ of a FBPNN trained by an improved FSDM, from (5), (35), and (42), we obtain:

$$\begin{cases} \hat{F}(k) = \sum_{j=1}^{\psi^M} (q_j - \beta_j)^2 = 0 \\ {}_i \rho_j^M(k) = D_{g_j^i}^1 \hat{F}(k) = -2(q_j - \beta_j)\left(D_{g_j^i}^1 \beta_j\right) = 0 \end{cases}, \quad (43)$$

Thus, from (34), when (43) is set up, we have:

$$\begin{cases} \hat{F}(k) = 0 \\ {}_i \rho_i^m(k) = D_{g_i^i}^1 \hat{F}(k) = 0 \end{cases}, \quad (44)$$

where $m = 0,1,\cdots,M-1$, $D_{w_{ij}^m}^1 \hat{F}(k) = {}_i \rho_i^m(k)\beta_j^{m-1} = \left[D_{b_i^m}^1 \hat{F}(k)\right]\beta_j^{m-1}$, and $D_{b_i^m}^1 \hat{F}(k) = {}_i \rho_i^m(k) = D_{b_i^m}^1 \hat{F}(k)$. Note that in (41), (43), and (44), for $\hat{F}(k) \geq 0$, thus when $\hat{F}(k) = 0$, $\beta_i$ is constant. Thus, if $\hat{F}(k) = 0$, we obtain $D_{g_i^i}^1 \beta_i = 0$ and $_i \rho_i^m(k) = 0$. Thus, in fact, in (41), (43) and (44), the second formula can be derived from the first one, but not vice versa. As we know, when and only when (43) and (44) are set up, the point of convergence of the iterative search processes (the final convergence result) is the same as the global minimum point of the square error $\hat{F}(k)$ of a FBPNN, namely, the iterative search algorithms in (27) and (28) can be guaranteed to converge to a fractional-order optimal minimum point $\left(w_{i,j}^{m*}, b_i^{m*}\right)$, a global minimum point, of the square error $\hat{F}(k)$ of a FBPNN trained by an improved FSDM. This completes the proof.

Lemma 1 indicates that at first, in (43) and (44), $D_{w_{ij}^m}^1 \hat{F}(k) = {}_i \rho_i^m(k)\beta_j^{m-1}$ and $D_{b_i^m}^1 \hat{F}(k) = {}_i \rho_i^m(k)$ merely consider the local characteristics of the square error $\hat{F}(k)$ of a classic first-order BPNN. Thus, a classic first-order BPNN is likely to converge to a local extreme point of its square error $\hat{F}(k)$. Assume that the iterative search process of a classic first-order BPNN converges to a local extreme point $\left(w_{i,j}^{m\prime}, b_i^{m\prime}\right) \neq \left(w_{i,j}^{m*}, b_i^{m*}\right)$. Thus, on this local extreme point $\left(w_{i,j}^{m\prime}, b_i^{m\prime}\right)$, one can obtain:

$$\begin{cases} \hat{F}(k) \neq 0 \\ {}_i \rho_i^m(k) = D_{g_i^i}^1 \hat{F}(k) = 0 \end{cases}, \quad (45)$$

However, in (25) and (26), $\widetilde{D}_{w_{ij}^m}^\nu \hat{F}(k) = \frac{\left[w_{ij}^m(k) - w_{int}\right]^{-\nu}}{\Gamma(1-\nu)}\hat{F}(k) + \sum_{n=1}^v \binom{v}{n}\frac{\left[w_{ij}^m(k) - w_{int}\right]^{n-\nu}}{\Gamma(n-\nu+1)} {}_n \rho_i^m(k)\beta_j^{m-1})^n$ and $\widetilde{D}_{b_i^m}^\nu \hat{F}(k) = \frac{\left[b_i^m(k) - b_{int}\right]^{-\nu}}{\Gamma(1-\nu)}\hat{F}(k) + \sum_{n=1}^v \binom{v}{n}\frac{\left[b_i^m(k) - b_{int}\right]^{n-\nu}}{\Gamma(n-\nu+1)} {}_n \rho_i^m(k)$ consider the nonlocal characteristics and the weak singularity of the square error $\hat{F}(k)$ of a FBPNN trained by an improved FSDM. Thus, a FBPNN trained by an improved FSDM can be guaranteed to converge to a global minimum point, of its square error $\hat{F}(k)$. The more efficient optimal searching capability of the fractional-order multi-scale global optimization of a FBPNN trained by an improved FSDM to determine the global optimal solution is the major advantage being superior to a classic first-order BPNN. Secondly, if the square error $\hat{F}(k)$ is mixed with white noise, it should increase the convergence time of the reverse incremental search of a FBPNN trained by an improved FSDM, which could cause (41) to be not set up. In this case, (41) could be invalid and hence the FBPNN may not converge to the fractional-order optimal point. Thirdly, the computational complexity of an algorithm can be typically measured by the number of its multiplications and additions, and related memory space, thus the computational complexity of a BPNN is linear with $\mathbf{w}^m = \left[w_{ij}^m\right]_{\psi^m \times \psi^{m-1}}$ and $\mathbf{b}^m = \left[b_i^m\right]_{\psi^m \times 1}$ ($i = 1 \to \psi^m$ and $j = 1 \to \psi^{m-1}$), which is in direct proportion to $O\!\left[\mathbf{w}^m\right] + O\!\left[\mathbf{b}^m\right]$ ($O[\ ]$ is the number of free parameters of a matrix). Therefore, for $n = 1,2,3$ in (27) and (28), the computational complexity of a FBPNN trained by an improved FSDM is in direct proportion to $4\left\{O\!\left[\mathbf{w}^m\right] + O\!\left[\mathbf{b}^m\right]\right\}$. Therefore, with the same number of neurons, the computational complexity of a FBPNN trained by an improved FSDM is 4 times greater than that of a classic first-order BPNN. In particular, Equations (27), (28), and (41) show that for the final convergence result of a FBPNN trained by an improved FSDM with $n=1$ is the same as that with $n = 1,2,3$. Thus, to further simplify the calculation of (27) and (28), without loss of generality, in the actual computations of a FBPNN trained by an improved FSDM, we can only set $n=1$. Then, in this case, with the same number of neurons, the computational complexity of a FBPNN trained by an improved FSDM is 2 times greater than that of a classic first-order BPNN. Fourthly, in general, the fractional-order extreme value of a normalized quadratic energy norm determined by the fractional-order partial derivatives is not equal to its integer-order one [40]. However, (3), (5), (41), and (44) indicate that if the global minimum value of the square error $\hat{F}(k)$ is equal to zero, the fractional-order optimal minimum point $\left(w_{i,j}^{m*}, b_i^{m*}\right)$ determined by the approximate fractional-order partial derivatives expressed by (25) and (26) is identical to the global minimum value of the square error $\hat{F}(k)$.

### 3.4 Assumption of structure of improved FSDM based FBPNN

In this subsection, an assumption of the structure of a FBPNN trained by an improved FSDM is made.

At first, in a similar manner to a classic first-order BPNN, a FBPNN trained by an improved FSDM behaviour is also highly dependent on the number of neurons and that of hidden layers. The aforementioned properties highlighted for a FBPNN trained by an improved FSDM regarding the fractional-order multi-scale global optimization search assume the sizing of a FBPNN trained by an improved FSDM is sufficient such that at least a fractional-order minimum point (a global minimum point) exists. With regards to an undersized improved FSDM based FBPNN, it does not exit a zero square error $\hat{F}(k)$. Thus, it is possible that the domain of attraction of the fractional-order optimal minimum point of an undersized improved FSDM based FBPNN does not contain the attractor $\left(w_{i,j}^{m*}, b_i^{m*}\right)$, or this domain of attraction is not included in the state space of an undersized improved FSDM based FBPNN at all. Assume that there is a local extreme point $\left(w_{i,j}^{m\prime}, b_i^{m\prime}\right)$ different from the fractional-order optimal minimum point $\left(w_{i,j}^{m*}, b_i^{m*}\right)$ of the square error $\hat{r}(k)$. Thus, on this local extreme point $\left(w_{c,j}^{m^l}, b_l^{m^l}\right)$, substitution of (45) into (27) and (28) results in:

$$\begin{cases} \left|w_{c,j}^m(k+1)-w_{c,j}^m(k)\right|=\left|w_{c,j}^m(k)-w_{c,j}^{m^l}\right|\neq 0 \\ \left|b_l^m(k+1)-b_l^m(k)\right|=\left|b_l^m(k)-b_l^{m^l}\right|\neq 0 \end{cases}. \tag{46}$$

Equation (46) indicates that the iterative search algorithms in (27) and (28) on a local extreme point $\left(w_{c,j}^{m^l}, b_l^{m^l}\right)$ cannot be ultimately terminated by themselves. On the one hand, if the domain of attraction of point $\left(w_{c,j}^{m^l}, b_l^{m^l}\right)$ is not entire and point $\left(w_{c,j}^{m^l}, b_l^{m^l}\right)$ is only in this domain of attraction, the iterative search process of an undersized improved FSDM based FBPNN should ultimately somewhat oscillate around point $\left(w_{c,j}^{m^l}, b_l^{m^l}\right)$ and attempt to identify point $\left(w_{c,j}^{m^*}, b_l^{m^*}\right)$. Otherwise, it should deviate from point $\left(w_{c,j}^{m^l}, b_l^{m^l}\right)$ to further determine point $\left(w_{c,j}^{m^*}, b_l^{m^*}\right)$. Conversely, if the domain of attraction of point $\left(w_{c,j}^{m^*}, b_l^{m^*}\right)$ is not included in the state space at all, the iterative search process of an undersized improved FSDM based FBPNN should deviate from point $\left(w_{c,j}^{m^l}, b_l^{m^l}\right)$ to make an continual attempt to search for point $\left(w_{c,j}^{m^*}, b_l^{m^*}\right)$.

Secondly, regarding a multilayer perceptron, Cybenko first mathematically proved that arbitrary decision regions can be arbitrarily well approximated by continuous feedforward neural networks with only a single hidden layer and any continuous sigmoidal nonlinearity [46]. Sontag derived a general result demonstrating that nonlinear control systems can be stabilized using two hidden layers; however, not in general using only a single hidden layer in a classic first-order BPNN [47]. Barron examined how the approximation error is related to the number of nodes in a classic first-order BPNN [48]. In a similar manner to a first-order BPNN, the approximation properties related to the number of neurons of the hidden layer in a FBPNN trained by an improved FSDM must be established. For the convenience of discussion, without loss of generality, assume that the actual output function $\beta_{\psi^l \times 1}^l(\boldsymbol{p}_{R\times 1})$ on $\boldsymbol{R}^{R\times 1}$ formulated in (3) is approximated by a FBPNN trained by an improved FSDM with a single layer of sigmoidal units, given as:

$$\beta_{\psi^l \times 1}^l(\boldsymbol{p}_{R\times 1})=\boldsymbol{f}^l\!\left(\boldsymbol{W}_{\psi^l \times R}^l \boldsymbol{p}_{R\times 1}+\boldsymbol{b}_{\psi^l \times 1}^l\right), \tag{47}$$

where $R$ denotes the dimensionality of the input matrix, $\psi^l \geq 1$ denotes the number of neurons of the hidden layer, and $\boldsymbol{f}^l$ is an arbitrary fixed sigmoidal function. Assume that $\beta_{\psi^l \times 1}^*(\boldsymbol{p}_{R\times 1})$ is a target output function of a FBPNN trained by an improved FSDM. If $\beta_{\psi^l \times 1}^*(\boldsymbol{p}_{R\times 1})$ is a causal signal with fractional primitives zero, from (1), we can derive the Fourier transform of the $v$-order derivative of $\beta_{\psi^l \times 1}^*(\boldsymbol{p}_{R\times 1})$, given as:

$$\mathrm{FT}\!\left[D_{\boldsymbol{p}_{R\times 1}}^v \beta_{\psi^l \times 1}^*(\boldsymbol{p}_{R\times 1})\right]=(\varsigma \boldsymbol{\varpi}_{R\times 1})^v \beta_{\psi^l \times 1}^*(\boldsymbol{\varpi}_{R\times 1}), \tag{48}$$

where $\mathrm{FT}[\ ]$ denotes the Fourier transform, $\varsigma$ denotes an imaginary unit, $\boldsymbol{\varpi}_{R\times 1}$ denotes angular frequency, and

$\beta_{\psi^l \times 1}^*(\boldsymbol{\varpi}_{R\times 1})=\iint_{\boldsymbol{R}^{R\times 1}}\beta_{\psi^l \times 1}^*(\boldsymbol{p}_{R\times 1})e^{-\varsigma \boldsymbol{\varpi}_{R\times 1}\boldsymbol{p}_{R\times 1}}d\boldsymbol{p}_{R\times 1}$ is the Fourier transform of $\beta_{\psi^l \times 1}^*(\boldsymbol{p}_{R\times 1})$. Thus, from (48), the smoothness property of $\beta_{\psi^l \times 1}^*(\boldsymbol{p}_{R\times 1})$ can be measured by the integrability of $\mathrm{FT}\!\left[D_{\boldsymbol{p}_{R\times 1}}^v \beta_{\psi^l \times 1}^*(\boldsymbol{p}_{R\times 1})\right]$ on $\boldsymbol{\varpi}^{R\times 1}$, given as:

$$C_\beta^v=\iint_{\boldsymbol{\varpi}^{R\times 1}}\left|\mathrm{FT}\!\left[D_{\boldsymbol{p}_{R\times 1}}^v \beta_{\psi^l \times 1}^*(\boldsymbol{p}_{R\times 1})\right]\right|d\boldsymbol{\varpi}_{R\times 1}=\iint_{\boldsymbol{\varpi}^{R\times 1}}\left|(\boldsymbol{\varpi}_{R\times 1})^v\right|\left|\beta_{\psi^l \times 1}^*(\boldsymbol{\varpi}_{R\times 1})\right|d\boldsymbol{\varpi}_{R\times 1}, \tag{49}$$

where $\left|(\boldsymbol{\varpi}_{R\times 1})^v\right|=\left[(\boldsymbol{\varpi}_{R\times 1})^v \bullet (\boldsymbol{\varpi}_{R\times 1})^v\right]^{1/2}$ and $\bullet$ denotes the inner product. Assume that the approximation error of a FBPNN trained by an improved FSDM can be measured by the integrated squared error regarding a probability $\rho$ on the hyper-ball $B_r=\{\boldsymbol{p}_{R\times 1}:\|\boldsymbol{p}_{R\times 1}\|\leq r\}$ of radius $r=\sup\|\boldsymbol{p}_{R\times 1}\|>0$. Therefore, we have the following lemma.

**Lemma 2**: If a target output function $\beta_{\psi^l \times 1}^*(\boldsymbol{p}_{R\times 1})$ with $C_\beta^v \geq 0$ finite is in the closure of the convex hull of a set $\boldsymbol{G}$ in a Hilbert space, with $\|\boldsymbol{g}\|<b$ for each $\boldsymbol{g}\in\boldsymbol{G}$ and $\bar{C}_\beta^v=\left(2rC_\beta^v\right)^2 \geq b^2-\left\|\beta_{\psi^l \times 1}^*(\boldsymbol{p}_{R\times 1})\right\|^2 \geq 0$, there should be an actual output function $\beta_{\psi^l \times 1}^l(\boldsymbol{p}_{R\times 1})$ of a FBPNN trained by an improved FSDM formulated in (47) in the convex hull of $\psi^l \geq 1$ points in $\boldsymbol{G}$, such that:

$$\iint_{B_r}\left[\beta_{\psi^l \times 1}^*(\boldsymbol{p}_{R\times 1})-\beta_{\psi^l \times 1}^l(\boldsymbol{p}_{R\times 1})\right]^2 \rho\, d\boldsymbol{p}_{R\times 1}\leq\frac{\bar{C}_\beta^v}{\psi^l}, \tag{50}$$

where $\|\ \|$ denotes the norm of the Hilbert space. If $\beta_{\psi^l \times 1}^l(\boldsymbol{p}_{R\times 1})$ is observed at sites $\{_i\boldsymbol{p}_{R\times 1}\}_{i=1\rightarrow S}$ restricted to $B_r$ that obey the uniform distribution, (50) provides a restricted condition on the approximation error, given as:

$$\frac{1}{S}\sum_{i=1}^S\left\|\beta_{\psi^l \times 1}^*(_i\boldsymbol{p}_{R\times 1})-\beta_{\psi^l \times 1}^l(_i\boldsymbol{p}_{R\times 1})\right\|^2 \leq\frac{\bar{C}_\beta^v}{\psi^l}, \tag{51}$$

where $S$ is the sample size.

**Proof**: We prove Lemma 2 based on the law of large numbers. Assume that $\beta_{\psi^l \times 1}^o(_i\boldsymbol{p}_{R\times 1})$ is a point in the convex hull of $\boldsymbol{G}$ being extremely close to $\beta_{\psi^l \times 1}^*(_i\boldsymbol{p}_{R\times 1})$, with $\left\|\beta_{\psi^l \times 1}^*(_i\boldsymbol{p}_{R\times 1})-\beta_{\psi^l \times 1}^o(_i\boldsymbol{p}_{R\times 1})\right\|\leq\delta/\psi^l$ and $\delta\geq 0$. $\boldsymbol{g}$ is stochastically drawn from the set $\{\boldsymbol{g}_k^o\in\boldsymbol{G}\}_{k=1\rightarrow m}$ with probability $P(\boldsymbol{g}=\boldsymbol{g}_k^o)=\rho_k\geq 0$. Then, if $m$ is sufficiently large, $\beta_{\psi^l \times 1}^o(_i\boldsymbol{p}_{R\times 1})=\sum_{k=1}^m \rho_k \boldsymbol{g}_k^o$ with $\sum_{k=1}^m \rho_k=1$. Further, assume that $\{\boldsymbol{g}_i^l\}_{i=1\rightarrow \psi^l}$ is independently drawn from the same distribution as $\boldsymbol{g}$. For the uniform distribution, $\rho=\rho_k=1/\psi^l$. If $\psi^l$ is sufficiently large, $\beta_{\psi^l \times 1}^l(_i\boldsymbol{p}_{R\times 1})=1/\psi^l\sum_{i=1}^{\psi^l}\boldsymbol{g}_i^l$ is a sample average. Thus, $E\left[\beta_{\psi^l \times 1}^l(_i\boldsymbol{p}_{R\times 1})\right]=\beta_{\psi^l \times 1}^o(_i\boldsymbol{p}_{R\times 1})$, where $E[\ ]$ denotes the mathematical expectation. Therefore, $E\left[\left\|\beta_{\psi^l \times 1}^l(_i\boldsymbol{p}_{R\times 1})-\beta_{\psi^l \times 1}^o(_i\boldsymbol{p}_{R\times 1})\right\|^2\right]=E\left[\left\|\beta_{\psi^l \times 1}^l(_i\boldsymbol{p}_{R\times 1})-\beta_{\psi^l \times 1}^o(_i\boldsymbol{p}_{R\times 1})\right\|^2\right]=\frac{1}{\psi^l}E\left[\left\|\boldsymbol{g}-\beta_{\psi^l \times 1}^o(_i\boldsymbol{p}_{R\times 1})\right\|^2\right]$, which equals $(1/\psi^l)\left[E\|\boldsymbol{g}\|^2-\left\|\beta_{\psi^l \times 1}^o(_i\boldsymbol{p}_{R\times 1})\right\|^2\right]$ that is restricted by $(1/\psi^l)\left[b^2-\left\|\beta_{\psi^l \times 1}^o(_i\boldsymbol{p}_{R\times 1})\right\|^2\right]$. For the mathematical expectation to be bounded in this manner, there must be $\{g_k^0\}_{k=1-sm}$, for which

$$\frac{1}{N}\sum_{i=1}^{N}\left\|\boldsymbol{\beta}_{\psi^1+1}^{\nu}(,\boldsymbol{p}_{R+1})-\boldsymbol{\beta}_{\psi^1+1}^{\nu}(,\boldsymbol{p}_{R+1})\right\|^2 \leq \frac{1}{\psi^1}\left[b^2-\left\|\boldsymbol{\beta}_{\psi^1+1}^{\nu}(\boldsymbol{p}_{R+1})\right\|^2\right].$$ Because

$\left\|\boldsymbol{\beta}_{\psi^1,1}^{\nu}(,\boldsymbol{p}_{R+1})-\boldsymbol{\beta}_{\psi^1+1}^{\nu}(,\boldsymbol{p}_{R+1})\right\| \leq \delta/\psi^1$ and $\overline{C_{\boldsymbol{\beta}}^{\nu}} > b^2 - \left\|\boldsymbol{\beta}_{\psi^1+1}^{\nu}(\boldsymbol{p}_{R+1})\right\|^2$, the proof of Lemma 2 is completed by the choice of a sufficiently small $\delta$.

Furthermore, let $\overline{C_{\boldsymbol{\beta}}^{\nu}} = (C_{\boldsymbol{\beta}}^{\nu})^2$ and add a statistically estimated regularization item $-\frac{R\nu^1}{S}\log S$ [48] to the restricted condition on the approximation error of a FBPNN trained by an improved FSDM formulated in (51), given as:

$$\frac{1}{S}\sum_{i=1}^{S}\left\|\boldsymbol{\beta}_{\psi^1+1}^{\nu}(,\boldsymbol{p}_{R+1})-\boldsymbol{\beta}_{\psi^1+1}^{\nu}(,\boldsymbol{p}_{R+1})\right\|^2 \leq \frac{(C_{\boldsymbol{\beta}}^{\nu})^2}{\psi^1}-\frac{R\nu^1}{S}\log S, \quad (52)$$

where $R$ is the number of input nodes. Equation (52) indicates that if the approximation error $\frac{1}{S}\sum_{i=1}^{S}\left\|\boldsymbol{\beta}_{\psi^1+1}^{\nu}(,\boldsymbol{p}_{R+1})-\boldsymbol{\beta}_{\psi^1+1}^{\nu}(,\boldsymbol{p}_{R+1})\right\|^2 = 0$, the number of neurons of the hidden layer in a FBPNN trained by an improved FSDM can be derived as:

$$\psi^1 = C_{\boldsymbol{\beta}}^{\nu}\left(\frac{S}{R\log S}\right)^{1/2}. \quad (53)$$

Equation (53) indicates that when the number of neurons of the hidden layer $\psi^1 = C_{\boldsymbol{\beta}}^{\nu}\left(\frac{S}{R\log S}\right)^{1/2}$, the sizing of a FBPNN trained by an improved FSDM is sufficient such that at least one fractional-order minimum exists.

### 3.5 *Fractional-order multi-scale global optimization of improved FSDM based FBPNN*

In this subsection, the fractional-order multi-scale global optimization of a FBPNN trained by an improved FSDM is analysed.

Equations (25)–(28) indicate that the variation of the fractional-order $\nu$ of the $\widetilde{D}_{w_{i,j}^\nu}^\nu\hat{F}$ and $\widetilde{D}_{b_i^\nu}^\nu\hat{F}$ can actually nonlinearly influence the learning process of a FBPNN trained by an improved FSDM to a certain degree. We restrict the square error $\hat{F}(k)$ to be a nonlinear increasing function of the fractional-order $\nu(k)$. Further, to simplify the calculation, we restrict $0 \leq \nu < 2$.

At first, (5), (25), and (26) indicate that on a given local extreme point and in its neighbourhood, we have $0 < \sigma_L^2 \leq \hat{F}(k) = e^2(k) < \sigma_U^2$, $\widetilde{D}_{w_{i,j}^\nu}^\nu\hat{F} \neq 0$ and $\widetilde{D}_{b_i^\nu}^\nu\hat{F} \neq 0$, where $\sigma_L$ and $\sigma_U$ are a sufficiently small positive scalar and a sufficiently large positive scalar, respectively. To escape from a local extreme point and its neighbourhood, the iterative search process of a FBPNN trained by an improved FSDM should have climbing capacity on $\hat{F}(k)$ by itself, i.e., $\widetilde{D}_{w_{i,j}^\nu}^\nu\hat{F}$ and $\widetilde{D}_{b_i^\nu}^\nu\hat{F}$ should be in nearly opposite directions of $D_{w_{i,j}^\nu}^1\hat{F}$ and $D_{b_i^\nu}^1\hat{F}$, respectively. On the one hand, if $D_{w_{i,j}^\nu}^1\hat{F} > 0$ or $D_{b_i^\nu}^1\hat{F} > 0$, we should restrict $\widetilde{D}_{w_{i,j}^\nu}^\nu\hat{F} < 0$ or $\widetilde{D}_{b_i^\nu}^\nu\hat{F} < 0$ correspondingly. Thus, from (25)–(28), we can derive that:

$$\widetilde{D}_{w_{i,j}^\nu}^\nu\hat{F}(k) = \frac{\left[w_{i,j}^m(k)-w_{inf}\right]^{-\nu}}{\Gamma(1-\nu)}\hat{F}(k) + \sum_{n=1}^{3}\binom{\nu}{n}\frac{\left[w_{i,j}^m(k)-w_{inf}\right]^{n-\nu}}{\Gamma(n-\nu+1)}{}_a\rho_j^m(k)(\beta_j^{m-1})^n < 0, \quad (54)$$

$$\widetilde{D}_{b_i^\nu}^\nu\hat{F}(k) = \frac{\left[b_i^m(k)-b_{inf}\right]^{-\nu}}{\Gamma(1-\nu)}\hat{F}(k) + \sum_{n=1}^{3}\binom{\nu}{n}\frac{\left[b_i^m(k)-b_{inf}\right]^{n-\nu}}{\Gamma(n-\nu+1)}{}_a\rho_i^m(k) < 0. \quad (55)$$

If we restrict $0 < \nu < 1$ in this case, we have $0 < 1/\Gamma(1-\nu) < 1$. Thus, if $0 < \nu < 1$, (54) and (55) can be simplified as:

$$\hat{F}(k) < -\Gamma(1-\nu)\sum_{n=1}^{3}\binom{\nu}{n}\frac{\left[w_{i,j}^m(k)-w_{inf}\right]^n}{\Gamma(n-\nu+1)}{}_a\rho_i^m(k)(\beta_j^{m-1})^n, \quad (56)$$

$$\hat{F}(k) < -\Gamma(1-\nu)\sum_{n=1}^{3}\binom{\nu}{n}\frac{\left[b_i^m(k)-b_{inf}\right]^n}{\Gamma(n-\nu+1)}{}_a\rho_i^m(k). \quad (57)$$

For $0 < \sigma_L^2 \leq \hat{F}(k) = e^2(k) < \sigma_U^2$, to enable (56) and (57) be set up, a necessary condition is given as:

$$0 < -\Gamma(1-\nu)\sum_{n=1}^{3}\binom{\nu}{n}\frac{\left[w_{i,j}^m(k)-w_{inf}\right]^n}{\Gamma(n-\nu+1)}{}_a\rho_i^m(k)(\beta_j^{m-1})^n = \Gamma(1-\nu)\left|\sum_{n=1}^{3}\binom{\nu}{n}\frac{\left[w_{i,j}^m(k)-w_{inf}\right]^n}{\Gamma(n-\nu+1)}{}_a\rho_i^m(k)(\beta_j^{m-1})^n\right|, \quad (58)$$

$$0 < -\Gamma(1-\nu)\sum_{n=1}^{3}\binom{\nu}{n}\frac{\left[b_i^m(k)-b_{inf}\right]^n}{\Gamma(n-\nu+1)}{}_a\rho_i^m(k) = \Gamma(1-\nu)\left|\sum_{n=1}^{3}\binom{\nu}{n}\frac{\left[b_i^m(k)-b_{inf}\right]^n}{\Gamma(n-\nu+1)}{}_a\rho_i^m(k)\right|. \quad (59)$$

Further, if $0 < \nu < 1$, $\binom{\nu}{n}\frac{1}{\Gamma(n-\nu+1)}\Big|_{n=1,2,3}$ obtains its maximum value when $n=1$. Thus, if $0 < \nu < 1$, from (58) and (59), according to the properties of inequality, (56) and (57) can be further simplified as:

$$\hat{F}(k) < \Gamma(1-\nu)\left|\sum_{n=1}^{3}\binom{\nu}{n}\frac{\left[w_{i,j}^m(k)-w_{inf}\right]^n}{\Gamma(n-\nu+1)}{}_a\rho_i^m(k)(\beta_j^{m-1})^n\right| \leq \frac{\nu}{1-\nu}\sum_{n=1}^{3}\left[w_{i,j}^m(k)-w_{inf}\right]^n{}_a\rho_i^m(k)(\beta_j^{m-1})^n = \sigma_1^2(\nu), \quad (60)$$

$$\hat{F}(k) < \Gamma(1-\nu)\left|\sum_{n=1}^{3}\binom{\nu}{n}\frac{\left[b_i^m(k)-b_{inf}\right]^n}{\Gamma(n-\nu+1)}{}_a\rho_i^m(k)\right| \leq \frac{\nu}{1-\nu}\sum_{n=1}^{3}\left[b_i^m(k)-b_{inf}\right]^n{}_a\rho_i^m(k) = \sigma_2^2(\nu). \quad (61)$$

From (60) and (61), one can further derive that:

$$0 < \nu_{T1} = 1/\left[\left(1/\hat{F}(k)\right)\sum_{n=1}^{3}\left[w_{i,j}^m(k)-w_{inf}\right]^n{}_a\rho_i^m(k)(\beta_j^{m-1})^n\right] + 1\right] < \nu < 1, \quad (62)$$

$$0 < \nu_{T2} = 1/\left[\left(1/\hat{F}(k)\right)\sum_{n=1}^{3}\left[b_i^m(k)-b_{inf}\right]^n{}_a\rho_i^m(k)\right] + 1\right] < \nu < 1. \quad (63)$$

In (60) and (62), $\sigma_L^2 = \min(\sigma_1^2(\nu), \sigma_1^2(\nu_{T1}))$ and $\sigma_U^2 = \max(\sigma_1^2(\nu), \sigma_1^2(\nu_{T1}))$. In (61) and (63), $\sigma_L^2 = \min(\sigma_2^2(\nu), \sigma_2^2(\nu_{T1}))$ and $\sigma_U^2 = \max(\sigma_2^2(\nu), \sigma_2^2(\nu_{T1}))$. In addition, if we restrict $1 < \nu < 2$ in this case, we have $-0.3 < 1/\Gamma(1-\nu) < 0$. Thus, if $1 < \nu < 2$, (54) and (55) can be simplified as:

$$\hat{F}(k) > -\Gamma(1-\nu)\sum_{n=1}^{3}\binom{\nu}{n}\frac{\left[w_{i,j}^m(k)-w_{inf}\right]^n}{\Gamma(n-\nu+1)}{}_a\rho_i^m(k)(\beta_j^{m-1})^n, \quad (64)$$

$$\hat{F}(k) > -\Gamma(1-\nu)\sum_{n=1}^{3}\binom{\nu}{n}\frac{\left[b_i^m(k)-b_{inf}\right]^n}{\Gamma(n-\nu+1)}{}_a\rho_i^m(k). \quad (65)$$

For $0 < \sigma_L^2 \leq \hat{F}(k) = e^2(k) < \sigma_U^2$, to enable (54) and (55) be set up, a weak restrictions is given as:

$$-\Gamma(1-\nu)\sum_{n=1}^{3}\binom{\nu}{n}\frac{\left[w_{i,j}^m(k)-w_{inf}\right]^n}{\Gamma(n-\nu+1)}{}_a\rho_i^m(k)(\beta_j^{m-1})^n\right| = -\Gamma(1-\nu)\sum_{n=1}^{3}\binom{\nu}{n}\frac{\left[w_{i,j}^m(k)-w_{inf}\right]^n}{\Gamma(n-\nu+1)}{}_a\rho_i^m(k)(\beta_j^{m-1})^n > 0, \quad (66)$$

$$-\Gamma(1-\nu)\left|\sum_{n=1}^{3}\binom{\nu}{n}\frac{\left[b_i^m(k)-b_{inf}\right]^n}{\Gamma(n-\nu+1)}{}_a\rho_i^m(k)\right| = -\Gamma(1-\nu)\sum_{n=1}^{3}\binom{\nu}{n}\frac{\left[b_i^m(k)-b_{inf}\right]^n}{\Gamma(n-\nu+1)}{}_a\rho_i^m(k) > 0. \quad (67)$$

Further, if $1 < \nu < 2$, $\binom{\nu}{n}\frac{1}{\Gamma(n-\nu+1)}\Big|_{n=1,2,3}$ also obtains its maximum value when $n=1$. Thus, if $1 < \nu < 2$, from (66) and (67), (64) and (65) can be further simplified as:

$$\hat{F}(k) > \sigma_L^2(v) = \frac{-v}{1-v} \sum_{n=1}^{3} \left[ w_{i,j}^m(k) - w_{int} \right]^n {}_a \rho_i^m(k) \left( \beta_j^{m-1} \right)^n \Big| \ge -\Gamma(1-v) \sum_{n=1}^{3} \left| \frac{\binom{v}{n} \left[ w_{i,j}^m(k) - w_{int} \right]^{n-v}}{\Gamma(n-v+1)} {}_a \rho_i^m(k) \left( \beta_j^{m-1} \right)^n \right|, \quad (68)$$

$$\hat{F}(k) > \sigma_L^2(v) = \frac{-v}{1-v} \sum_{n=1}^{3} \left[ b_j^m(k) - b_{int} \right]^n {}_a \rho_i^m(k) \Big| \ge -\Gamma(1-v) \sum_{n=1}^{3} \left| \Gamma(1-v) \frac{\binom{v}{n} \left[ b_j^m(k) - b_{int} \right]^{n-v}}{\Gamma(n-v+1)} {}_a \rho_i^m(k) \right|, \quad (69)$$

From (68) and (69), one can further derive that:

$$1 < v_{T3} = -1 / \left[ \left[ \left( 1 / \hat{F}(k) \right) \sum_{n=1}^{3} \left[ w_{i,j}^m(k) - w_{int} \right]^n {}_a \rho_i^m(k) \left( \beta_j^{m-1} \right)^n \right| - 1 \right] < v < 2, \quad (70)$$

$$1 < v_{T4} = -1 / \left[ \left[ \left( 1 / \hat{F}(k) \right) \sum_{n=1}^{3} \left[ b_j^m(k) - b_{int} \right]^n {}_a \rho_i^m(k) \right] - 1 \right] < v < 2. \quad (71)$$

In (68) and (70), $\sigma_L^2 = \min(\sigma_L^2(v), \sigma_L^2(v_{T3}))$ and $\sigma_U^2 = \max(\sigma_L^2(v), \sigma_L^2(v_{T3}))$. In (69) and (71), $\sigma_L^2 = \min(\sigma_L^2(v), \sigma_L^2(v_{T4}))$ and $\sigma_U^2 = \max(\sigma_L^2(v), \sigma_L^2(v_{T4}))$. On the other hand, if $D_{w_{i,j}^m}^v \hat{F} < 0$ or $D_{b_j^m}^v \hat{F} < 0$, we should restrict $\tilde{D}_{w_{i,j}^m}^v \hat{F} > 0$ or $\tilde{D}_{b_j^m}^v \hat{F} > 0$ correspondingly. Thus, from (25)–(28), we can derive that:

$$\tilde{D}_{w_{i,j}^m}^v \hat{F}(k) = \frac{\left[ w_{i,j}^m(k) - w_{int} \right]^{-v}}{\Gamma(1-v)} \hat{F}(k) + \sum_{n=1}^{3} \frac{\binom{v}{n} \left[ w_{i,j}^m(k) - w_{int} \right]^{n-v}}{\Gamma(n-v+1)} {}_a \rho_i^m(k) \left( \beta_j^{m-1} \right)^n > 0, \quad (72)$$

$$\tilde{D}_{b_j^m}^v \hat{F}(k) = \frac{\left[ b_j^m(k) - b_{int} \right]^{-v}}{\Gamma(1-v)} \hat{F}(k) + \sum_{n=1}^{3} \frac{\binom{v}{n} \left[ b_j^m(k) - b_{int} \right]^{n-v}}{\Gamma(n-v+1)} {}_a \rho_i^m(k) > 0. \quad (73)$$

In a similar way, if we restrict $0 < v < 1$ in this case, the following is true:

$$0 < v < 1 / \left[ \left[ \left( 1 / \hat{F}(k) \right) \sum_{n=1}^{3} \left[ w_{i,j}^m(k) - w_{int} \right]^n {}_a \rho_i^m(k) \left( \beta_j^{m-1} \right)^n \right| + 1 \right] = v_{T1} < 1, \quad (74)$$

$$0 < v < 1 / \left[ \left[ \left( 1 / \hat{F}(k) \right) \sum_{n=1}^{3} \left[ b_j^m(k) - b_{int} \right]^n {}_a \rho_i^m(k) \right] + 1 \right] = v_{T2} < 1. \quad (75)$$

In addition, if we restrict $1 < v < 2$ in this case, the following is true:

$$1 < v < -1 / \left[ \left[ \left( 1 / \hat{F}(k) \right) \sum_{n=1}^{3} \left[ w_{i,j}^m(k) - w_{int} \right]^n {}_a \rho_i^m(k) \left( \beta_j^{m-1} \right)^n \right| - 1 \right] = v_{T3} < 2, \quad (76)$$

$$1 < v < -1 / \left[ \left[ \left( 1 / \hat{F}(k) \right) \sum_{n=1}^{3} \left[ b_j^m(k) - b_{int} \right]^n {}_a \rho_i^m(k) \right] - 1 \right] = v_{T4} < 2. \quad (77)$$

Secondly, when $0 \le \hat{F}(k) < \sigma_L^2$, the iterative search process of a FBPNN trained by an improved FSDM is in the neighbourhood of a fractional-order optimal minimum point. Equations (62), (63), (74), and (75) further show that to encourage the iterative search process of a FBPNN trained by an improved FSDM to converge to this given fractional-order optimal minimum point, the square error $\hat{F}(k)$ should decrease until it is equal to zero. Thus, when $0 \le \hat{F}(k) < \sigma_L^2$, if $D_{w_{i,j}^m}^v \hat{F} > 0$ or $D_{b_j^m}^v \hat{F} > 0$, we restrict $0 \le v < v_{T1} < 1$ for $w_{i,j}^m$ or $0 \le v < v_{T2} < 1$ for $b_j^m$ correspondingly. Note that from (1), $D_x^v \big|_{v=0}$ is an identity operator, which implements neither differential nor integral. Thus, we have $D_x^0 0 = 0$. Equations (22) and (23) restrict $v = 0$ on a fractional-order optimal minimum point with $\hat{F}(k) = 0$. When $v = 0$, the iterative search process of a FBPNN trained by an improved FSDM can be terminated on a fractional-order optimal minimum point of the square error $\hat{F}(k)$ by itself; If $D_{w_{i,j}^m}^v \hat{F} < 0$ or $D_{b_j^m}^v \hat{F} < 0$, we set $0 < v_{T1} < v \le 1$ for $w_{i,j}^m$ or $0 < v_{T2} < v \le 1$ for $b_j^m$ correspondingly.

Thirdly, when $\hat{F}(k) \ge \sigma_U^2$, we can directly set $v = 1$.

Note that at first, when $0 < \sigma_L^2 \le \hat{F}(k) < \sigma_U^2$, $\tilde{D}_{w_{i,j}^m}^v \hat{F}$ and $\tilde{D}_{b_j^m}^v \hat{F}$ are restricted in the near opposite directions of $D_{w_{i,j}^m}^1 \hat{F}$ and $D_{b_j^m}^1 \hat{F}$, respectively. As we know, the first-order gradient of a point is in the direction of the fastest growth of the scalar field and its module value is the maximum rate of change. In the numerical implementation, the directions of $D_{w_{i,j}^m}^1 \hat{F}$ and $D_{b_j^m}^1 \hat{F}$ are those of their maximum module values, respectively. For example, on the two-dimensional discrete plane, there are eight directional derivatives (their interval is 45 degrees) of a point $(w_{i,j}^m, b_j^m)$ in its neighbourhood. We select the maximum directional derivative of a point $(w_{i,j}^m, b_j^m)$ to be $D_{w_{i,j}^m}^1 \hat{F}$ and $D_{b_j^m}^1 \hat{F}$. Secondly, to enhance the multi-scale searching capability, we should also set the learning rate, $\mu(k)$, of a FBPNN trained by an improved FSDM to be an appropriate correlation function with $\tilde{D}_{w_{i,j}^m}^v \hat{F}$, $\tilde{D}_{b_j^m}^v \hat{F}$, and $\hat{F}(k)$. Thirdly, to guarantee to escape the domain of attraction of a given local extreme point, as long as $\tilde{D}_{w_{i,j}^m}^v \hat{F}$ and $\tilde{D}_{b_j^m}^v \hat{F}$ begin to climb, we should keep climbing until them get to the top of the hill in the neighbourhood of this local extreme point. On the top of the hill, we should restrict the downward direction to be different from the previous uphill direction of $\tilde{D}_{w_{i,j}^m}^v \hat{F}$ and $\tilde{D}_{b_j^m}^v \hat{F}$.

## 4. Experiment and Analysis

### 4.1 *Example function approximation of improved FSDM based FBPNN*

In this subsection, for the following Examples 1–4, we discuss an example of the function approximation of a FBPNN trained by an improved FSDM.

As we know, multilayer networks can be used to approximate virtually any function if we have a sufficient number of neurons in the hidden layers. We choose a network for a FBPNN trained by an improved FSDM and apply it to a particular problem. Without loss of generality, we illustrate the characteristics with a $1 - \psi^1 - 1$ improved FSDM based FBPNN, which is displayed in Fig. 2.

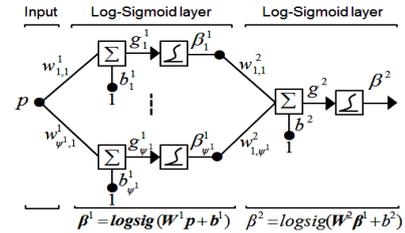

Fig. 2. Example of function approximation of a FBPNN trained by an improved FSDM.

In Fig. 2, $\psi^1$ denotes the number of neurons in the first layer (hidden layer) of a FBPNN trained by an improved FSDM. The activation functions for the first and second layer are log-sigmoid, which can be expressed as follows:

$$f^1(x) = f^2(x) = 1 / \left( 1 + e^{-x} \right). \quad (78)$$

To simplify the analysis, we assign a specific problem to a FBPNN trained by an improved FSDM. We know the global optimization solution to this problem. In Fig. 2, we set $\psi^1 = 2$. It is also assumed that the function to be approximated is the response of the same a 1-2-1 improved FSDM based FBPNN, with the following values for the weights and biases: $w_{1,1}^1 = 10$, $w_{2,1}^1 = 10$, $b_1^1 = -5$, $b_2^1 = 5$, $w_{1,1}^2 = 1$, $w_{1,2}^2 = 1$, and $b^2 = -1$. The response for these parameters is displayed in Fig. 3, which consists of the plot of this improved FSDM based FBPNN output $\rho^2$ as the input $p$ is varied over the range [-2,2].

The improved FSDM based FBPNN described in Fig. 2 must be trained to approximate the function displayed in Fig. 3. The approximation is exact when this improved FSDM based FBPNN parameters are set to the following values: $w_{1,1}^1 = 10$, $w_{2,1}^1 = 10$, $b_1^1 = -5$, $b_2^1 = 5$, $w_{1,1}^2 = 1$, $w_{1,2}^2 = 1$, and $b^2 = -1$. We assume that the function to be approximated is sampled at the values $p = -2, -1.9, -1.8, \cdots, 2$ with equal probability. The mean squared error of this improved FSDM based FBPNN is equal to the average sum of the squared errors at these 41 points. To plot the mean squared error of this improved FSDM based FBPNN in three-dimensional space, we vary only two parameters simultaneously. Figure 4 illustrates the mean squared error of this improved FSDM based FBPNN when only two parameters, $w_{1,1}^1$ and $w_{1,1}^2$, are adjusted; the other parameters are set to their aforementioned optimal values.

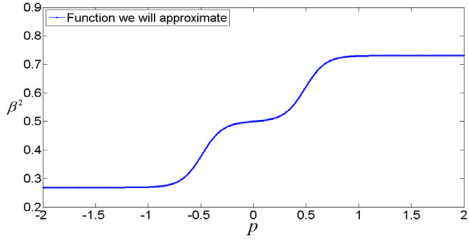

Fig. 3. Function to be approximated.

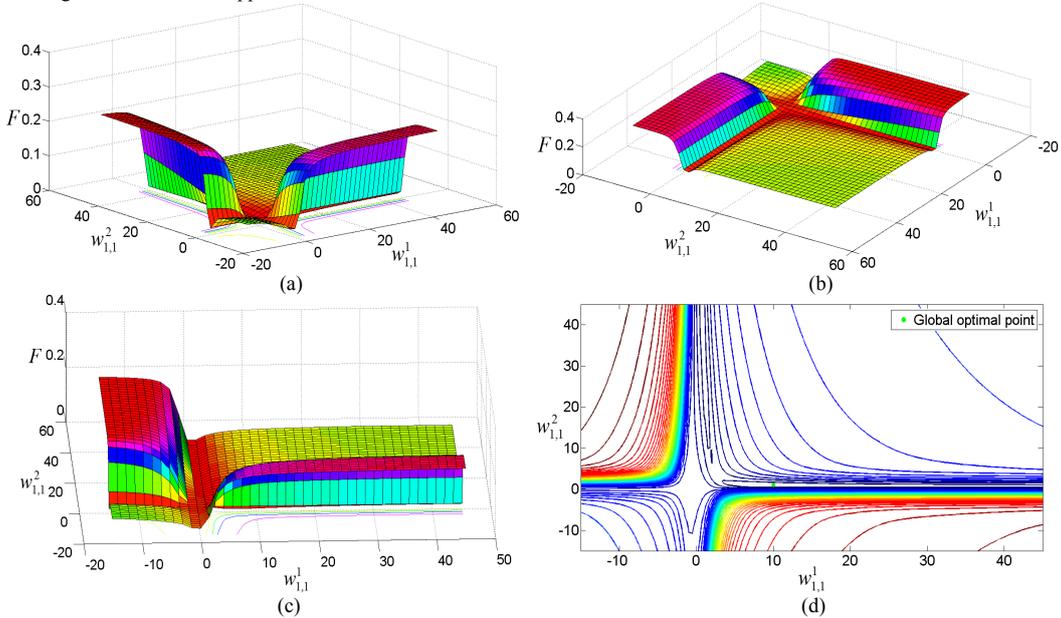

Fig. 4. Mean squared error of a FBPNN trained by an improved FSDM: (a) Front view of mean squared error; (b) Back view of mean squared error; (c) Lateral view of mean squared error; (d) Contour map of mean squared error.

It is observed that the optimal minimum error occurs when $w_{1,1}^1 = 10$ and $w_{1,1}^2 = 1$, as indicated by the solid green circle in Fig. 4.

### 4.2 Fractional-order multi-scale global optimization of improved FSDM based FBPNN

In this subsection, we analyse the fractional-order multi-scale global optimization of a FBPNN trained by an improved FSDM.

Inspired by the aforementioned mathematical analysis, the fractional-order $v(k)$ of $\widetilde{D}_{w_u^v} \hat{F}(k)$ and $\widetilde{D}_{b_i^v} \hat{F}(k)$ of a FBPNN trained by an improved FSDM can be self-adaptively determined. To simplify the multi-scale searching process in the following experiments, we construct an imperfect adaptive kernel function of the fractional-order $v$ at the $k^{th}$ iteration of a FBPNN trained by an improved FSDM as follows:

$$v(k) = 2 \left| \frac{1 - \Phi^{-e(k)}}{1 + \Phi^{-e(k)}} \right| + |e(k)|, \tag{79}$$

where $\Phi = \left| \rho^M \right|^{[2+e(k)]}$, $e(k) = (1/\psi^M) \sum_{i=1}^{\psi^M} e_i(k)$ is the average error at the $k^{th}$ iteration, and $\rho^M = (1/\psi^M) \sum_{i=1}^{\psi^M} \rho_i^M$ is the first-order average sensibility in the $M^{th}$ layer, the last layer, of a FBPNN trained by an improved FSDM.

From (3), (5), (10), (24), and (78), we can derive the first-order average sensibility in the output layer of the improved FSDM based FBPNN in Fig. 2, $_1\rho_i^2(k)\big|_{i=1\sim n_l}=\rho_i^2(k)=-2e(k)\beta_i^2\left(1-\beta_i^2\right)$. The output $\beta_1^2=\beta_i^2\big|_{i=1\sim n_l}$ of a FBPNN trained by an improved FSDM changes dynamically with the initial condition, weight matrices, bias matrices, and average error $e(k)$. For the convenience of illustration, we analyse the multi-scale adjustment of the adaptive kernel function of $v(k)$ by assuming the output $\beta_1^2=\beta_i^2\big|_{i=1\sim n_l}=1+e(k)$, $\beta_1^1\big|_{j=1\sim n_2}=1$, $w_{i,j}^2(k)\big|_{i=1\sim n_l,\ j=1\sim n_2}=1$, and $b_i^2(k)\big|_{i=1\sim n_l}=1$ in (79), which is displayed in Fig. 5.

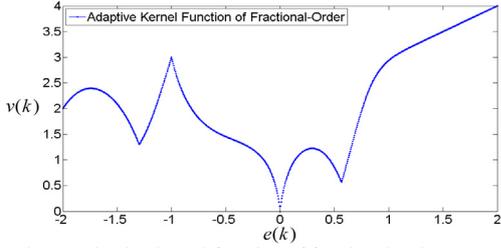

Fig. 5. Adaptive kernel function of fractional-order $v(k)$.

From Fig. 5 and (79), we can observe that the fractional-order $v(k)$ adaptively varies with the error $e(k)$ during the entire iterative search process of a FBPNN trained by an improved FSDM. As in the previous discussion, if the minimum value of the square error $\hat{F}(k)$ is equal to zero, the fractional order $v(k)$ approaches zero on a fractional-order optimal extreme point (global optimal minimum point) with zero average error $e(k)=0$.

***Example 1***: In this example, we select an initial condition in the specific zone from a random sample of cases, where the reverse incremental search of a FBPNN trained by an improved FSDM and a classic first-order BPNN can both converge to the global optimal minimum point of the square error $\hat{F}(k)$ at the $k^{th}$ iteration. We set the same parameters, the rate of convergence $\mu=5.50$ and the number of iterations $=2000$, for both a FBPNN trained by an improved FSDM and a first-order BPNN in this simulation experiment; the initial condition is at the point where $w_{1,1}^1=-4$ and $w_{1,1}^2=-4$. From (27)–(30), (78), and (79), the iterative search process of a FBPNN trained by an improved FSDM and a first-order BPNN can be represented as displayed in Fig. 6.

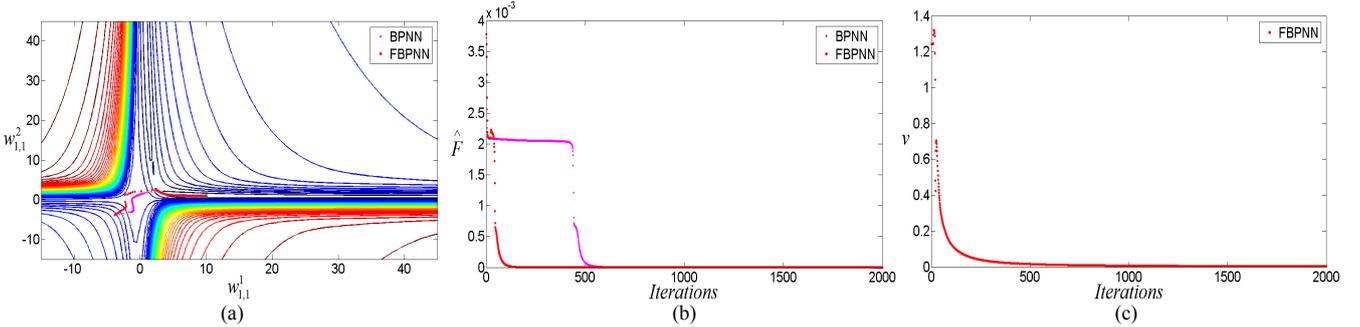

(a)       (b)       (c)

Fig. 6. Comparison of iterative search processes of a FBPNN trained by an improved FSDM and a first-order BPNN: (a) Convergence trajectories; (b) Convergence patterns of squared error of $k^{th}$ iteration; (c) Fractional-order $v$ of FBPNN.

In Fig. 6(a), we observe two convergence trajectories of a FBPNN trained by an improved FSDM and a first-order BPNN (batch mode) when only two parameters, $w_{1,1}^1$ and $w_{1,1}^2$, are varied. Figures 6(a) and 6(b) indicate that regarding the aforementioned initial condition ($w_{1,1}^1=-4$ and $w_{1,1}^2=-4$), a FBPNN trained by an improved FSDM and a first-order BPNN eventually converge to an optimal point ($w_{1,1}^1=10$ and $w_{1,1}^2=1$). The square errors $\hat{F}(k)$ of a FBPNN trained by an improved FSDM and a first-order BPNN on an optimal point are equal to zero. The two convergence trajectories of a FBPNN trained by an improved FSDM and a first-order BPNN bypass an initial flat surface and then fall into a gently sloping valley, as observed in Fig. 6(a). From Fig. 6(a), we can observe that a FBPNN trained by an improved FSDM requires a fewer number of iterations than a first-order BPNN to converge to the optimal point. However, as per the aforementioned discussion, with the same number of neurons, the computational complexity of a FBPNN trained by an improved FSDM is 4 times greater than a first-order BPNN in every iterative computation. A FBPNN trained by an improved FSDM requires additional computational time compared to a first-order BPNN. Thus, a first-order BPNN is faster to converge to the optimal point than a FBPNN trained by an improved FSDM. Figure 6(c) indicates that as the value of $\hat{F}(k)$ changes over the convergence trajectory of a FBPNN trained by an improved FSDM, the fractional-order $v(k)$ of a FBPNN trained by an improved FSDM varies according to the adaptive kernel function in (79). Thus, the fractional-order $v(k)$ of a FBPNN trained by an improved FSDM in its entire iterative search process is not constant, which demonstrates the fractional-order multi-scale global optimization of a FBPNN trained by an improved FSDM.

***Example 2***: In this example, we select an initial condition in the gently sloping zone of the square error $\hat{F}(k)$ from a random sample of cases. We set the same parameters, the rate of convergence $\mu = 5.50$, the number of iterations = 9000, for both a FBPNN trained by an improved FSDM and a first-order BPNN in this simulation experiment; the initial condition is at the point where $w^1_{1,1} = 5$ and $w^2_{1,1} = 30$. From (27)–(30), the iterative search process of a FBPNN trained by an improved FSDM and a first-order BPNN can be represented as displayed in Fig. 7.

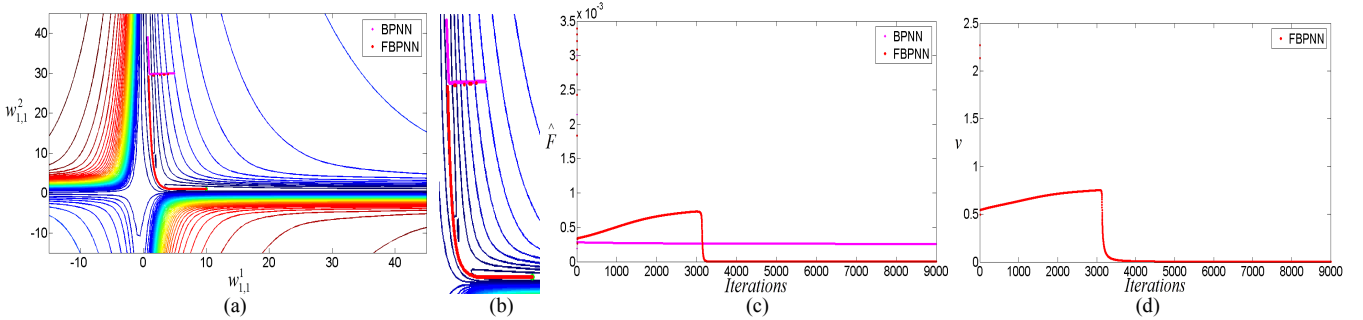

Fig. 7. Comparison of iterative search process of a FBPNN trained by an improved FSDM and a first-order BPNN: (a) Convergence trajectories; (b) Local magnification for convergence trajectories; (c) Convergence patterns of squared error of $k^{th}$ iteration; (d) Fractional-order $v$ of FBPNN.

In Figs. 7(a) and (b), the convergence trajectory of a first-order BPNN illustrates the manner in which it can converge to a local extreme point ( $w^1_{1,1} = 0.7003$ and $w^2_{1,1} = 35.2626$ ). The convergence trajectory of a first-order BPNN is trapped in a valley and diverges from the global optimal solution. However, the convergence trajectory of a FBPNN trained by an improved FSDM illustrates the manner in which it can converge to a global optimal minimum point ( $w^1_{1,1} = 10$ and $w^2_{1,1} = 1$ ). The convergence trajectory of a FBPNN trained by an improved FSDM passes over a valley and converges to the global optimal solution. Figure 7(c) indicates that when the number of iterations of a first-order BPNN is approximately equal to 6500, the square error $\hat{F}(k)$ of a first-order BPNN reduces to a nonzero minimum, which is approximately equal to $0.25 \cdot 10^{-3}$. As the number of iterations of a first-order BPNN increases, this nonzero minimum cannot continue to reduce. Thus, a first-order BPNN can be trapped in a local extreme point. Figure 7(c) also indicates that when the number of iterations of a FBPNN trained by an improved FSDM increases from one to ten, its square error $\hat{F}(k)$ decreases sharply from $3.4 \cdot 10^{-3}$ to $0.35 \cdot 10^{-3}$. When the number of iterations of a FBPNN trained by an improved FSDM increases from 11 to 3000, its square error $\hat{F}(k)$ does not decrease; rather, it increases gradually from $0.35 \cdot 10^{-3}$ to $0.82 \cdot 10^{-3}$. Further, when the number of iterations of a FBPNN trained by an improved FSDM increases from 3001 to 4000, its square error $\hat{F}(k)$ decreases sharply from $0.82 \cdot 10^{-3}$ to a zero minimum. As the number of iterations of a FBPNN trained by an improved FSDM increases, this minimum remains at zero. Thus, a FBPNN trained by an improved FSDM converges to a global optimal minimum point. From (1), (27), (28), and (79), we observe that when $0 < v < 1$, the square error

$\hat{F}(k)$ of a FBPNN trained by an improved FSDM may not always decrease; it can increase at certain points, which helps a FBPNN trained by an improved FSDM to bypass the first-order local extreme points of $\hat{F}(k)$. Figure 7(d) indicates that when the number of iterations of a FBPNN trained by an improved FSDM increases from one to 3000, its fractional-order $v$ decreases sharply from 2.30 to 0.51. When the number of iterations of a FBPNN trained by an improved FSDM increases from 11 to 3000, its fractional-order $v$ increases gradually from 0.51 to 0.80. Further, when the number of iterations of a FBPNN trained by an improved FSDM increases from 3001 to 4000, its fractional-order $v$ decreases sharply from 0.80 to zero. From (79), we can observe that the fractional-order $v$ of a FBPNN trained by an improved FSDM varies with its average error $e(k)$ changing during the entire iterative process. Thus, the fractional-order $v$ of a FBPNN trained by an improved FSDM in its entire iterative search process is not constant, which demonstrates the fractional-order multi-scale global optimization of a FBPNN trained by an improved FSDM.

***Example 3***: To further verify the fractional-order multi-scale global optimization of a FBPNN trained by an improved FSDM, we select another initial condition in the sharply sloping zone of the square error $\hat{F}(k)$ from a random sample of cases. We set the same parameters, the rate of convergence $\mu = 5.50$, the number of iterations = 6000, for both a FBPNN trained by an improved FSDM and a first-order BPNN in this simulation experiment; the initial condition is at the point $w^1_{1,1} = -8$ and $w^2_{1,1} = 9$. From (27)–(30), the iterative search process of a FBPNN trained by an improved FSDM and a first-order BPNN can be represented as displayed in Fig. 8.

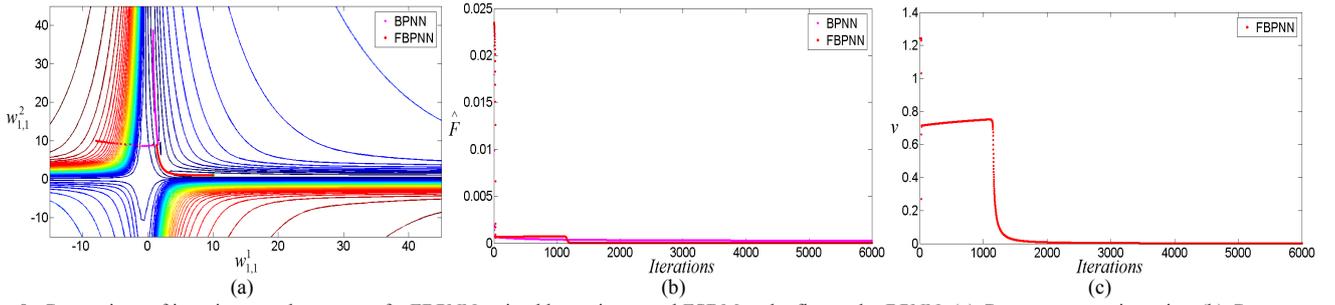

Fig. 8. Comparison of iterative search process of a FBPNN trained by an improved FSDM and a first-order BPNN: (a) Convergence trajectories; (b) Convergence patterns of squared error of $k^{th}$ iteration; (c) Fractional-order $\nu$ of FBPNN.

In Fig. 8(a), the convergence trajectory of a first-order BPNN illustrates the manner in which it can converge to a local extreme point ($w^1_{1,1} = 0.7003$ and $w^2_{1,1} = 35.2626$). However, the convergence trajectory of a FBPNN trained by an improved FSDM illustrates the manner in which it can converge to a global optimal minimum point ($w^1_{1,1} = 10$ and $w^2_{1,1} = 1$). Figure 8(b) indicates that when the number of iterations of a first-order BPNN is approximately equal to 2000, the square error $\hat{F}(k)$ of a first-order BPNN reduces to a nonzero minimum, which is approximately equal to $0.25 \cdot 10^{-3}$. Thus, a first-order BPNN can be trapped in a local extreme point. However, when the number of iterations of a FBPNN trained by an improved FSDM is approximately equal to 2300, the square error $\hat{F}(k)$ of a FBPNN trained by an improved FSDM reduces to a zero minimum. Thus, a FBPNN trained by an improved FSDM converges to a global optimal minimum point. Figure 8(b) also indicates that when $0 < \nu < 1$, the square error $\hat{F}(k)$ of a FBPNN trained by an improved FSDM may not always decrease; rather, it can increase at certain points, which helps a FBPNN trained by an improved FSDM to bypass the first-order local extreme points of $\hat{F}(k)$. Figure 8(c) indicates that the fractional-order $\nu$ of a FBPNN trained by an improved FSDM in its entire iterative search process is not constant; rather, it varies according to (79), which demonstrates the fractional-order multi-scale global optimization of a FBPNN trained by an improved FSDM.

***Example 4***: We present an extreme example where the initial condition is directly on a local extreme point of the square error $\hat{F}(k)$. Assume we have set the same parameters, the rate of convergence $\mu = 5.50$, the number of iterations = 9000, for both a FBPNN trained by an improved FSDM and a first-order BPNN in this simulation experiment, and the initial condition is directly at the local extreme point ($w^1_{1,1} = 0.7003$ and $w^2_{1,1} = 35.2626$) of $\hat{F}(k)$. Thus, from (27)–(30), the iterative search process of a FBPNN trained by an improved FSDM and a first-order BPNN can be represented as displayed in Fig. 9.

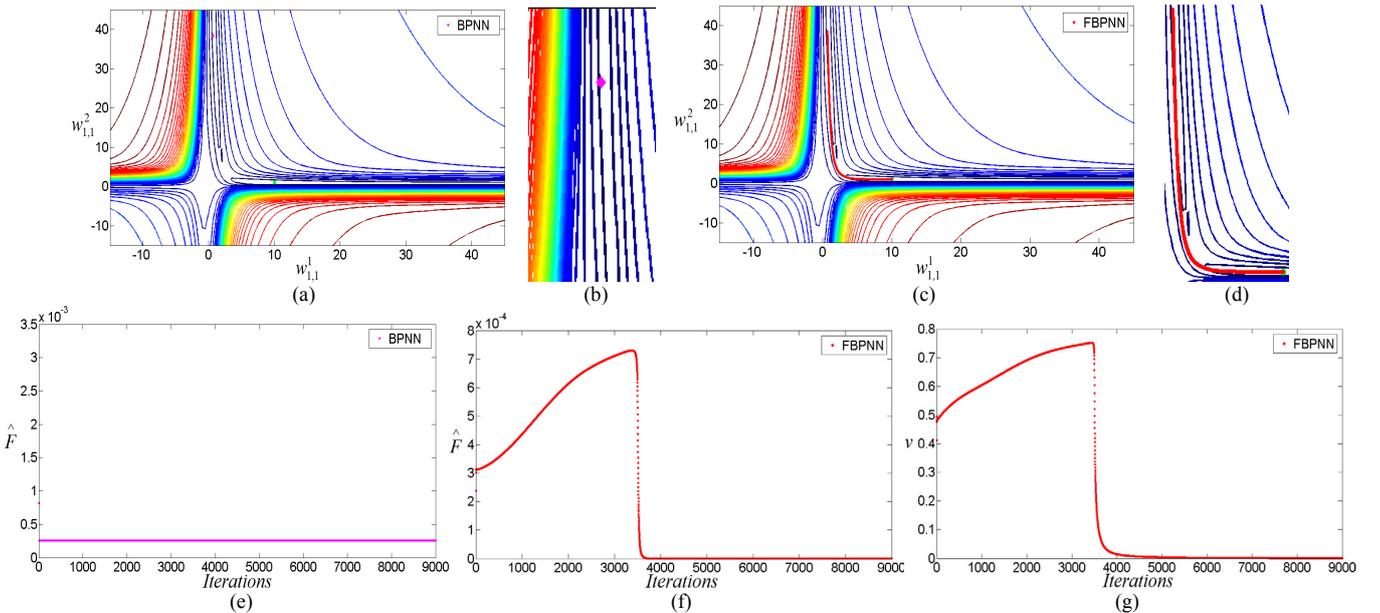

Fig. 9. Comparison of iterative search process of a FBPNN trained by an improved FSDM and a first-order BPNN: (a) Convergence trajectory of BPNN; (b) Local magnification for (a); (c) Convergence trajectory of FBPNN; (d) Local magnification for (c); (e) Convergence patterns of squared error of $k^{th}$ iteration of BPNN; (f) Convergence patterns of squared error of $k^{th}$ iteration of FBPNN; (g) Fractional-order $\nu$ of FBPNN.

Figures 9(a) and 9(e) indicate that if the initial condition is directly at a local extreme point of $\hat{F}(k)$, the convergence trajectory of a first-order BPNN is trapped at this local extreme point and the square error $\hat{F}(k)$ remains unchanged. Figure 9(c) indicates that even if the initial condition is directly at a local extreme point of $\hat{F}(k)$, the convergence trajectory of a FBPNN trained by an improved FSDM can converge to a global optimal minimum point ($w_{1,1}^1 = 10$ and $w_{1,1}^2 = 1$). Figure 9(f) indicates that the number of iterations of a FBPNN trained by an improved FSDM increases from one to 3500, its $\hat{F}(k)$ does not decrease; rather it increases gradually from $0.25 \cdot 10^{-3}$ to $7.4 \cdot 10^{-4}$. Further, when the number of iterations of a FBPNN trained by an improved FSDM increases from 3501 to 5200, the $\hat{F}(k)$ decreases sharply from $7.4 \cdot 10^{-4}$ to a zero minimum. As the number of iterations of a FBPNN trained by an improved FSDM increases, this zero minimum remains at zero. Thus, a FBPNN trained by an improved FSDM converges to a global optimal minimum point. Figure 9(g) indicates that the fractional-order $v$ of a FBPNN trained by an improved FSDM in its entire iterative search process is not constant; rather, it varies according to (79), which demonstrates the fractional-order multi-scale global optimization of a FBPNN trained by an improved FSDM.

### 4.3 Comparative performance of improved FSDM based FBPNN with real data

In this subsection, to further verify the fractional-order multi-scale global optimization, we implement two comparative performances of a FBPNN trained by an improved FSDM with real data.

***Example 5***: In this example, the data presented in Table 1 displays the output as a transfer function of the input of a nonlinear signal-processing filter.

Table 1. Input and output of a nonlinear signal-processing filter.

| Input | −1 | − 0.9 | − 0.8 | − 0.7 | − 0.6 | − 0.5 | − 0.4 | − 0.3 | − 0.2 | − 0.1 |
|---|---|---|---|---|---|---|---|---|---|---|
| Output | −0.832 | −0.423 | −0.024 | 0.344 | 1.282 | 3.456 | 4.020 | 3.232 | 2.102 | 1.504 |
| Input | 0 | 0.1 | 0.2 | 0.3 | 0.4 | 0.5 | 0.6 | 0.7 | 0.8 | 0.9 |
| Output | 0.248 | 1.242 | 2.344 | 3.262 | 2.052 | 1.684 | 1.022 | 2.224 | 3.022 | 1.984 |

We choose a FBPNN trained by an improved FSDM and a Levenberg-Marquardt algorithm-based first-order BPNN [17] to provide two nonlinear least-square approximations to the data set in Table 1. We illustrate the characteristics with a $1-\psi^1-1$ improved FSDM based FBPNN and a same structural first-order BPNN. The number of neurons in the first layer (hidden layer) of a FBPNN trained by an improved FSDM and a first-order BPNN are both equal to $\psi^1 = 15$. The activation functions for the first layer and second layer are tan-sigmoid $f^1(x) = \left[2/\left(1 + e^{-2x}\right)\right] - 1$ and pure linear function $f^2(x) = x$, respectively. $\boldsymbol{W}_{15\times1}^1 = [$ $w_{1,1}^1$ , 20.8597, 21.2543, -21.0232, -21.3975, 21.0826, -21.0743, 21.0052, 21.0272, 20.9446, $w_{11,1}^1$, -21.1307, 21.2419, 20.9357, 21.0157$]^T$ and $\boldsymbol{W}_{1\times15}^2 = [-0.7629,$ -0.7168, 1.1592, 0.4330, 0.9470, 0.5903, -1.1983, -0.7002, -0.3756, -1.0144, -0.2451, -1.3834, 0.4546, 0.2460, 0.3230] are randomly selected as the weight matrices of the first and second layer of a FBPNN trained by an improved FSDM and a first-order BPNN, respectively. $\boldsymbol{b}_{15\times1}^1 = [-21.0070, -18.1627,$ -14.6449, 11.9684, 8.0087, -5.7329, 2.0816, 0.7399, 2.7071, 6.1967, -8.9802, -11.7774, 14.6532, 18.0707, 20.9846$]^T$ and $\boldsymbol{b}_{1\times1}^2 = [-0.4954]$ are also randomly selected as the bias matrices of the first and second layer of a FBPNN trained by an improved FSDM and a first-order BPNN, respectively. For the convenience of illustration, we simultaneously vary only two parameters of a FBPNN trained by an improved FSDM. Figure 10 illustrates the mean squared error of the FBPNN when only two parameters, $w_{1,1}^1$ and $w_{11,1}^1$, are adjusted; the other parameters are set to their aforementioned randomly selected values.

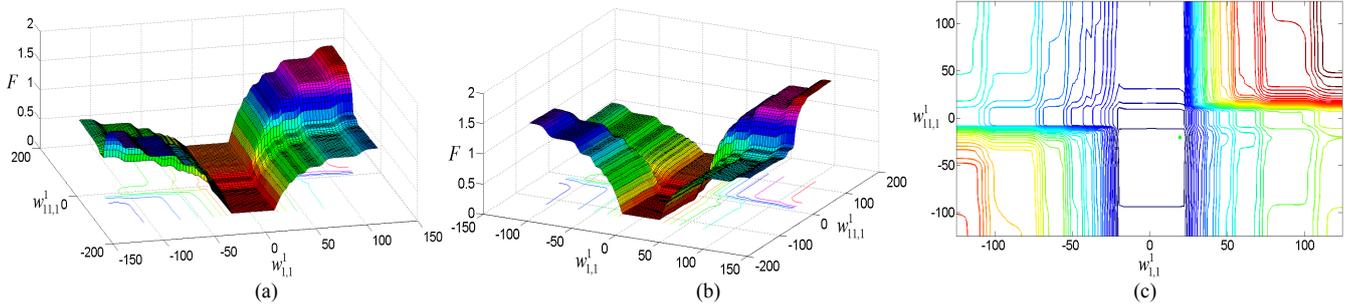

Fig. 10. Mean squared error of a FBPNN trained by an improved FSDM: (a) Left side view of mean squared error; (b) Right side view of mean squared error; (c) Contour map of mean squared error.

It can be observed that a relative optimal minimum error occurs when $w_{1,1}^1 = 19.3065$ and $w_{11,1}^1 = -20.4575$, as indicated by the solid green circle in Fig. 10. We set the same parameters, the rate of convergence $\mu = 3.50$, the number of iterations is equal to 3000, for both a FBPNN trained by an improved FSDM and a first-order BPNN in this simulation experiment and the initial conditions were at the points where ($w_{1,1}^1 = 108$ and $w_{11,1}^1 = 116$), ($w_{1,1}^1 = -110$ and $w_{11,1}^1 = -106$), and ($w_{1,1}^1 = -95$ and $w_{11,1}^1 = 100$). From (27)–(30), the iterative search process of a FBPNN trained by an improved FSDM and a first-order BPNN can be represented as in Fig. 11.

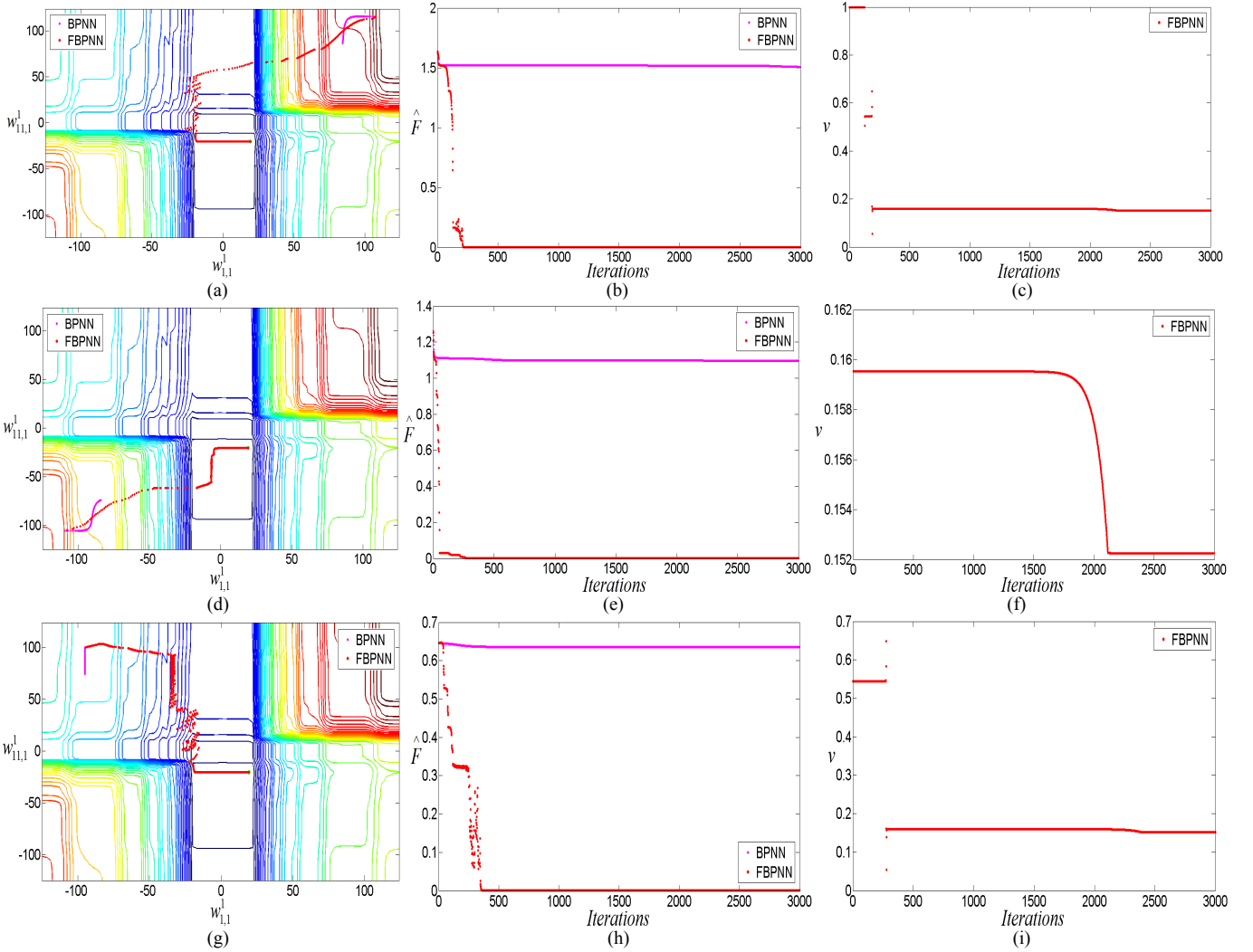

Fig. 11. Comparison of iterative search process of a FBPNN trained by an improved FSDM and a first-order BPNN: (a) Convergence trajectories ( $w_{1,1}^1 = 108$ and $w_{11,1}^1 = 116$ ); (b) Convergence patterns of squared error of $k^{th}$ iteration ( $w_{1,1}^1 = 108$ and $w_{11,1}^1 = 116$ ); (c) Fractional-order $v$ of FBPNN ( $w_{1,1}^1 = 108$ and $w_{11,1}^1 = 116$ ); (d) Convergence trajectories ( $w_{1,1}^1 = -110$ and $w_{11,1}^1 = -106$ ); (e) Convergence patterns of squared error of $k^{th}$ iteration ( $w_{1,1}^1 = -110$ and $w_{11,1}^1 = -106$ ); (f) Fractional-order $v$ of FBPNN ( $w_{1,1}^1 = -110$ and $w_{11,1}^1 = -106$ ); (g) Convergence trajectories ( $w_{1,1}^1 = -95$ and $w_{11,1}^1 = 100$ ); (h) Convergence patterns of squared error of $k^{th}$ iteration ( $w_{1,1}^1 = -95$ and $w_{11,1}^1 = 100$ ); (i) Fractional-order $v$ of FBPNN ( $w_{1,1}^1 = -95$ and $w_{11,1}^1 = 100$ ).

In Figs. 11(a), (d), and (g), the convergence trajectories illustrate that a first-order BPNN can converge to three local extreme points, ( $w_{1,1}^1 = 86.0196$ and $w_{11,1}^1 = 84.169$ ), ( $w_{1,1}^1 = -83.8969$ and $w_{11,1}^1 = -74.0297$ ), and ( $w_{1,1}^1 = -94.9789$ and $w_{11,1}^1 = 74.0876$ ). However, all convergence trajectories of illustrate that a FBPNN trained by an improved FSDM can converge to a relative global optimal minimum point ( $w_{1,1}^1 = 19.3065$ and $w_{11,1}^1 = -20.4575$ ). Figures 11(b), (e), and (h) display that at first, when the number of iterations increases, the square error $\hat{F}(k)$ of a first-order BPNN reduces to a nonzero minimum, which is clearly greater than that of a FBPNN trained by an improved FSDM. Secondly, because we vary only two parameters ( $w_{1,1}^1$ and $w_{11,1}^1$ ) simultaneously, the maximum adjustment of a FBPNN trained by an improved FSDM converges to the relative global optimal minimum point ( $w_{1,1}^1 = 19.3065$ and $w_{11,1}^1 = -20.4575$ ), where its minimum square error $\hat{F}(k)$ ( $\hat{F}_{min} = 0.001115$ ) approaches zero, however, not equal to zero. If we vary all the parameters simultaneously, the convergence trajectory of a FBPNN trained by an improved FSDM can converge to a global optimal minimum point, where its minimum square error $\hat{F}(k)$ is equal to zero. Figures 11(c), (f), and (i), and (79) indicate that the minimum square error $\hat{F}(k)$ ( $\hat{F}_{min} = 0.001115$ ) approaches zero and the minimum of the fractional-order $v$ of a FBPNN trained by an improved FSDM approaches zero, however, not equal to zero.

In the following extreme example, the initial condition is directly on a local extreme point of the square error $\hat{F}(k)$. We set the same parameters, the rate of convergence $\mu = 3.50$, the number of iterations be equal to 3000 for both a FBPNN trained by an improved FSDM and a first-order BPNN in this simulation experiment; the initial condition is directly at a local extreme point ($w_{11,1}^1 = -9.00$ and $w_{11,1}^1 = 8.2676$) of $\hat{F}(k)$. Thus, from (27)–(30), the iterative search process of a FBPNN trained by an improved FSDM and a first-order BPNN can be represented as indicated in Fig. 12.

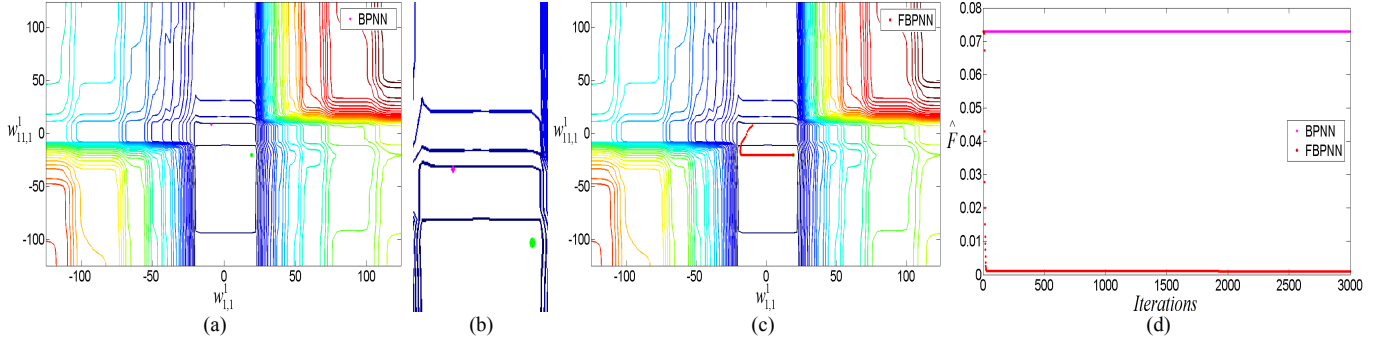

Fig. 12. Comparison of iterative search process of a FBPNN trained by an improved FSDM and a first-order BPNN: (a) Convergence trajectory of BPNN; (b) Local magnification for (a); (c) Convergence trajectory of FBPNN; (d) Convergence patterns of squared error of $k^{th}$ iteration.

Figure 12 indicates that if the initial condition is directly at a local extreme point of $\hat{F}(k)$, the convergence trajectory of a first-order BPNN are trapped at this local extreme point and the square error $\hat{F}(k)$ remains unchanged. Conversely, even if the initial condition is directly at a local extreme point of $\hat{F}(k)$, the convergence trajectory of a FBPNN trained by an improved

FSDM can converge to a relative global optimal minimum point ($w_{11,1}^1 = 19.3065$ and $w_{11,1}^1 = -20.4575$).

The responses of a FBPNN trained by an improved FSDM and a first-order BPNN for the convergence parameters of Fig. 11 are displayed in Fig. 13; this consists of the plots of the outputs of a FBPNN trained by an improved FSDM and a first-order BPNN as the inputs are varied over the range of Table 1.

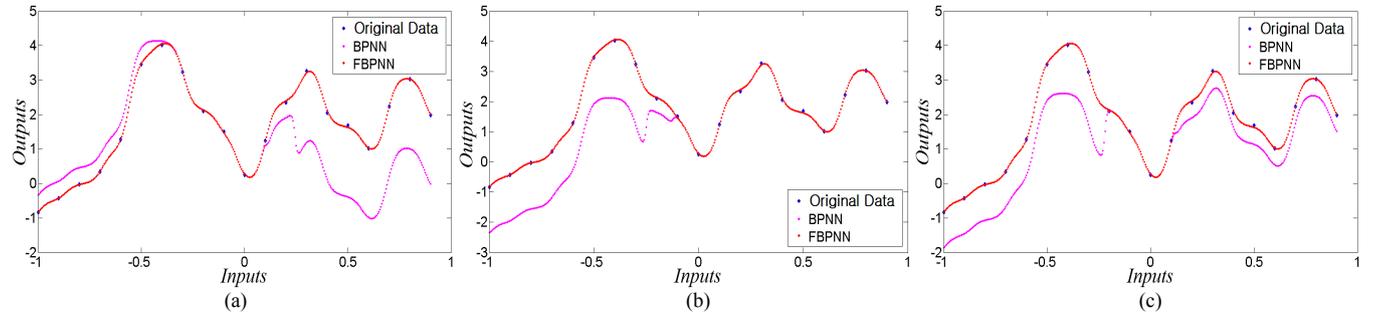

Fig. 13. Responses of a FBPNN trained by an improved FSDM and a first-order BPNN for the convergence parameters: (a) Convergence parameters of Fig. 11(a); (b) Convergence parameters of Fig. 11(d); (c) Convergence parameters of Fig. 11(g).

Figure 13 indicates that for the convergence parameters of Fig. 11, the sample data of Table 1 can be fit well by a FBPNN trained by an improved FSDM. The fitting error of a first-order BPNN is clearly greater than a FBPNN trained by an improved FSDM.

## 5. Conclusions

The application of fractional calculus to neural networks and cybernetics is an emerging field of study and only a small number of studies have been conducted in this area. The properties of the fractional calculus of a signal are considerably different from those of its integer-order calculus.

Fractional calculus has been applied to neural networks and cybernetics primarily owing to its inherent advantages of long-term memory, nonlocality, and weak singularity. Therefore, to improve the optimization performance of the ordinary a first-order BPNNs, it is logical to generalize a first-order BPNN to a FBPNN trained by an improved FSDM by applying a state-of-the-art application of promising mathematical method, fractional calculus. From this inspiration, in this study, a FBPNN trained by an improved FSDM was achieved whose reverse incremental search was in the negative directions of the approximate fractional-order partial derivatives of the square error $\hat{F}(k)$. The more efficient optimal searching capability of the fractional-order multi-scale global optimization of a FBPNN trained by an improved FSDM to determine the global optimal solution is the major advantage being superior to a classic first-order BPNN.

From the aforementioned discussion, we can also observe that there are other problems that must be further studied. For example, the aforementioned imperfect adaptive kernel function of the fractional-order $v$ at the $k^{th}$ iteration of a FBPNN trained by an improved FSDM, $v_{(k)}$, is not sufficient for an arbitrary quadratic energy norm $\hat{F}_{(k)}$. Further, the computational complexity of a FBPNN trained by an improved FSDM increases with the introduction of fractional calculus. With the same number of neurons, the computational complexity of a FBPNN trained by an improved FSDM is 4 times greater than that of a classic first-order BPNN in every iterative computation. Therefore, it is evident that other topics such as how to construct an efficient appropriate correlation function of $\mu_{(k)}$, how to construct a more efficient adaptive kernel function of the fractional-order $v_{(k)}$, and how to reduce the computational effort of a FBPNN trained by an improved FSDM must be studied further. These topics will be discussed in our future work.

### ACKNOWLEDGEMENTS

This work was supported by the National Key Research and Development Program Foundation of China under Grants 2018YFC0830300, and the National Natural Science Foundation of China under Grants 61571312.